\documentclass[pdflatex,iicol]{sn-jnl}
\usepackage{times}

\usepackage[utf8]{inputenc}
\usepackage[english]{babel}
\usepackage[autostyle,english=american]{csquotes}
\usepackage{microtype}
\usepackage{subcaption}
\usepackage{graphicx}%
\usepackage{multirow}%
\usepackage{amsmath,amssymb,amsfonts,amsthm}%
\usepackage{mathrsfs}%
\usepackage[title]{appendix}%
\usepackage{xcolor}%
\usepackage{textcomp}%
\usepackage{manyfoot}%
\usepackage{booktabs}%
\usepackage{algorithm}%
\usepackage{algorithmicx}%
\usepackage{algpseudocode}%
\usepackage{listings}%
\usepackage{glossaries}
\usepackage{array}
\usepackage{physics}
\usepackage{tabularx}
\usepackage{longtable}
\usepackage{makecell}
\usepackage{float}

\newcolumntype{Y}{>{\centering\arraybackslash}X}
\newacronym{sar}{SaR}{Search-and-Rescue}
\newacronym{mpc}{MPC}{Model Predictive Control}
\newacronym{tmpc}{TMPC}{Tube-based Model Predictive Control}
\newacronym{sddtmpc}{SDD-TMPC}{State-Dependent Dynamic Tube-based Model Predictive Control}
\newacronym{rl}{RL}{Reinforcement Learning}
\newacronym{hri}{HRI}{Human-Robot Interaction}
\newacronym{bfs}{BFS}{Breadth-First Search}
\newacronym{ssi}{SSI}{Sequential Single Item}
\newacronym{pomdp}{POMDP}{Partially Observable Markov Decision Process}
\newacronym{cmdp}{CMDP}{Constrained Markov Decision Process}
\newacronym{ppo}{PPO}{Proximal Policy Optimization}
\newacronym{lstm}{LSTM}{Long Short-Term Memory}
\newacronym{gae}{GAE}{Generalized Advantage Estimation}
\newacronym{mlp}{MLP}{Multilayer Perceptron}
\newacronym{bdi}{BDI}{Belief-Desire-Intention}
\newacronym{ci}{CI}{Confidence Interval}
\newacronym{pasar}{PA-SaR}{Perception-Aware Search-and-Rescue}

\newcommand{\bqtask}{\boldsymbol q_t^{\mathrm{target}}}
\newcommand{\bqtaskprime}{\boldsymbol q_{t^{\prime}}^{\mathrm{target}}}
\newcommand{\bqprime}{\boldsymbol q_{r^{\prime}}^{\mathrm{r}}}
\newcommand{\bq}{\boldsymbol{q}^{\mathrm{r}}_r}

\newcommand{\bqv}{\boldsymbol{q}^{\mathrm{v}}_v}
\newcommand{\bzvict}{\boldsymbol{z}^{\mathrm{victim}}}
\newcommand{\bzallvict}{\boldsymbol{Z}^{\mathrm{victim}}}
\newcommand{\zvict}{\boldsymbol{z}_v^{\mathrm{victim}}}
\newcommand{\bu}{\boldsymbol{u}_r}

\newcommand{\nh}{n^{\mathrm{h}}}
\newcommand{\nw}{n^{\mathrm{w}}}
\newcommand{\tburn}{T^{\mathrm{burn}}}
\newcommand{\ptrip}{p_r^{\mathrm{displace}}}
\newcommand{\sigmaaconf}{\sigma_{r}^{\mathrm{confidence}}}
\newcommand{\rrobot}{\rho^{\mathrm{r}}_r}
\newcommand{\rvict}{\rho^{\mathrm{v}}_v}
\newcommand{\alphatrip}{\alpha^{\mathrm{displace}}}
\newcommand{\betatrip}{\beta^{\mathrm{displace}}}
\newcommand{\qrobot}{Q^{\mathrm{r}}_r}
\newcommand{\qsetrobot}{\mathcal{Q}^{\mathrm{r}}}
\newcommand{\qvictim}{Q_v^{\mathrm{v}}}

\newcommand{\mtrav}{M^{\mathrm{traverse}}}
\newcommand{\mvuln}{M^{\mathrm{vulnerable}}}
\newcommand{\mijvuln}{M_{ij}^{\mathrm{vulnerable}}}
\newcommand{\mijprimevuln}{M_{i^{\prime}j^{\prime}}^{\mathrm{vulnerable}}}
\newcommand{\mburn}{M^{\mathrm{burn}}}
\newcommand{\mijburn}{M_{ij}^{\mathrm{burn}}}
\newcommand{\mfire}{M^{\mathrm{fire}}}
\newcommand{\mvict}{M^{\mathrm{victim}}}
\newcommand{\mrobot}{M^{\mathrm{robot}}}
\newcommand{\ptrav}{P_r^{\mathrm{r,traverse}}}
\newcommand{\probot}{P_r^{\mathrm{r,robot}}}
\newcommand{\pvuln}{P_r^{\mathrm{r,vulnerable}}}
\newcommand{\pijexpl}{P_{r,ij}^{\mathrm{r,exploration}}}
\newcommand{\pijvuln}{P_{r,ij}^{\mathrm{r,vulnerable}}}
\newcommand{\pijconf}{P_{r,ij}^{\mathrm{r,confidence}}}
\newcommand{\pijprimeconf}{P_{r,i^{\prime}j^{\prime}}^{\mathrm{r,confidence}}}
\newcommand{\pvict}{P_r^{\mathrm{r,victim}}}
\newcommand{\pfire}{P_r^{\mathrm{r,fire}}}
\newcommand{\pconf}{P_r^{\mathrm{r,confidence}}}
\newcommand{\pexpl}{P_r^{\mathrm{r,exploration}}}

\newcommand{\mijtrav}{M^{\mathrm{traverse}}_{ij}}
\newcommand{\nusafe}{\nu^{\mathrm{safe}}}
\newcommand{\numoderate}{\nu^{\mathrm{moderate}}}
\newcommand{\nuhigh}{\nu^{\mathrm{high}}}
\newcommand{\pignite}{p^{\mathrm{ignite}}}
\newcommand{\pignitehat}{\hat{p}^{\mathrm{ignite}}}
\newcommand{\deltavuln}{\delta_r^{\mathrm{vulnerable}}}
\newcommand{\tburnhat}{\hat{T}^{\mathrm{burn}}}
\newcommand{\wdisc}{w^{\mathrm{discover}}}
\newcommand{\wconf}{w^{\mathrm{confidence}}}
\newcommand{\wvuln}{w^{\mathrm{vulnerable}}}
\newcommand{\wterm}{w^{\mathrm{terminal}}}
\newcommand{\wdet}{w^{\mathrm{detected}}}
\newcommand{\wstep}{w^{\mathrm{step}}}
\newcommand{\wterr}{w^{\mathrm{collide}}}
\newcommand{\wvict}{w^{\mathrm{victim}}}
\newcommand{\wsucc}{w^{\mathrm{success}}}
\newcommand{\wcov}{w^{\mathrm{coverage}}}
\newcommand{\womega}{w^{\Omega}}
\newcommand{\wprox}{w^{\mathrm{proximity}}}
\newcommand{\wgain}{w^{\mathrm{gain}}}
\newcommand{\wloss}{w^{\mathrm{loss}}}
\newcommand{\omegaconftilde}{\tilde{\omega}^{\mathrm{confidence}}}
\newcommand{\omegavulntilde}{\tilde{\omega}^{\mathrm{vulnerable}}}
\newcommand{\omegaproxtilde}{\tilde{\omega}^{\mathrm{proximity}}}
\newcommand{\omegadisc}{\omega^{\mathrm{discover}}}
\newcommand{\omegaconf}{\omega^{\mathrm{confidence}}}
\newcommand{\omegapath}{\omega^{\mathrm{path}}}
\newcommand{\omegavuln}{\omega^{\mathrm{vulnerable}}}
\newcommand{\omegaprox}{\omega^{\mathrm{proximity}}}
\newcommand{\phidisc}{\Phi_r^{\mathrm{discover}}}
\newcommand{\phiconf}{\Phi_r^{\mathrm{confidence}}}
\newcommand{\phivuln}{\Phi_r^{\mathrm{vulnerable}}}
\newcommand{\phiprox}{\Phi_r^{\mathrm{proximity}}}
\newcommand{\phipath}{\Phi_r^{\mathrm{path}}}
\newcommand{\zdiscover}{Z_r^{\mathrm{discover}}}
\newcommand{\tasktype}{\tau^{\mathrm{type}}_r}
\newcommand{\taskstatus}{\tau^{\mathrm{status}}_r}

\theoremstyle{thmstyleone}%
\theoremstyle{thmstyletwo}%

\theoremstyle{thmstylethree}%

\begin{document}

\glsdisablehyper

\title[PA-SaR: Perception-Aware Search and Rescue]
{PA-SaR: Perception-Aware Learning, Control, and Coordination
for Multi-Robot Search and Rescue}

%%=============================================================%%
%% GivenName	-> \fnm{Joergen W.}
%% Particle	-> \spfx{van der} -> surname prefix
%% FamilyName	-> \sur{Ploeg}
%% Suffix	-> \sfx{IV}
%% \author*[1,2]{\fnm{Joergen W.} \spfx{van der} \sur{Ploeg} 
%%  \sfx{IV}}\email{iauthor@gmail.com}
%%=============================================================%%

\author[1]{\fnm{Jokūbas} \sur{Dimša}}
\equalcont{These authors contributed equally to this work.}

\author[1]{\fnm{Maria} \sur{Grouev}}
\equalcont{These authors contributed equally to this work.}

\author[1]{\fnm{Filip} \sur{Keerberg}}
\equalcont{These authors contributed equally to this work.}

\author[1]{\fnm{Nicolas} \sur{Laplagne}}
\equalcont{These authors contributed equally to this work.}

\author[1]{\fnm{Tigo} \sur{Schutgens}}
\equalcont{These authors contributed equally to this work.}

\author[1,2]{\fnm{Sven} \sur{Schoonebeek}}\email{s.schoonebeek@tudelft.nl}

\author[1]{\fnm{Anahita} \sur{Jamshidnejad}}\email{a.jamshidnejad@tudelft.nl}

\affil[1]{\orgdiv{Department of Intelligent Systems, Faculty of Electrical Engineering, Mathematics, and Computer Science}, \orgname{TU Delft}, \orgaddress{\street{Van Mourik Broekmanweg 6}, \city{Delft}, \postcode{2628 XE}, \state{Zuid-Holland}, \country{The Netherlands}}}

\affil[2]{\orgdiv{Delft Center for Systems and Control, Faculty of Mechanical Engineering, TU Delft}, \orgname{TU Delft}, \orgaddress{\street{Mekelweg 2}, \city{Delft}, \postcode{2628 CE}, \state{Zuid-Holland}, \country{The Netherlands}}}

\abstract{
Autonomous \gls{sar} requires robots to explore unknown or partially known environments, 
avoid hazards, coordinate team activities, and guide discovered victims to safety. 
Improving individual capabilities of robots does not necessarily improve mission outcomes: 
safer motion can restrict the exploration behavior, faster discovery can increase the rescue demand, 
and effective guidance depends on victim responses. 
We introduce \gls{pasar}, a mathematical and architectural framework for coordinating 
heterogeneous local control, learning-based exploration, supervisory task allocation, 
and adaptive approach-lead victim guidance around mission-level objectives. 
\gls{pasar} distinguishes environmental ground-truth, robot-specific perceptions, and simulated victim behavior. 
This provides a common basis for integrating complementary mechanisms 
and analyzing their interactions across mission stages. 
We demonstrate the framework through single- and multi-robot simulation studies 
incorporating model-predictive and learning-based exploration, perception-based safety shielding, 
event-driven task allocation, and adaptive victim guidance. 
The results show trade-offs between collision avoidance and search completion, 
indicate that supervisory allocation improves search completion beyond information sharing alone, 
and demonstrate that increased exploration does not necessarily yield more rescues. Adaptive guidance increases rescues relative to continuous leading, 
without a detected improvement in mean or final victim trust or a reduction in mortality. 
These findings support evaluating local control, team coordination, 
and evacuation as related but distinct aspects of autonomous \gls{sar} 
within the considered simulation environments.
}


%%\pacs[JEL Classification]{D8, H51}

%%\pacs[MSC Classification]{35A01, 65L10, 65L12, 65L20, 65L70}

\maketitle

\glsresetall
\section{Introduction}
\label{sec:introduction}

\gls{sar} missions are time-critical: the probability of finding
survivors decreases as the mission is prolonged \cite{b1}. This urgency,
combined with the danger faced by rescue workers, motivates the use of
robotic systems to extend their reach into otherwise inaccessible areas
while reducing their exposure to hazardous conditions \cite{b2}.
When these robots operate autonomously, mission safety becomes partly
dependent on decisions made from incomplete information. Unsafe
robot actions may injure victims, damage robots that are crucial for the mission, or block a route, while poor
victim guidance can leave a discovered victim stranded. Discovery alone
does not bring a victim to safety, and delaying rescue until exploration
is complete can postpone assistance to those victims. %already found. 
This
motivates coordinating continued exploration with rescue: explorers, tasked with mapping and victim detection, continue
searching while rescue-capable robots guide located victims toward an
exit. Coverage and discovery are therefore important intermediate
measures, but evacuation outcomes are also needed to assess mission
effectiveness.

Safe exploration requires local motion planning that accounts for
uncertain sensing and disturbances. \gls{mpc} optimizes predicted
motion subject to constraints \cite{b12}, while tube-based
\gls{mpc} tightens these constraints to accommodate disturbances
within a prescribed set \cite{b3}. A fixed disturbance bound can
exclude plans that would be feasible under smaller local
disturbances. \gls{sddtmpc} addresses this conservativeness by
adapting the tube to state-dependent disturbances and has been
evaluated in simulated \gls{sar} scenarios \cite{b4}. However, these model predictive controllers still require online optimization \cite{b13}, motivating complementary approaches to action selection.

A trained \gls{rl} policy can select actions without explicitly searching
over future control sequences at each decision step \cite{b14}.
Deep \gls{rl}, which combines \gls{rl} and deep learning, has been applied to maritime search coverage, prioritizing
likely victim locations while reducing repeated coverage \cite{b37}.
However, optimizing a reward does not itself ensure safe behavior:
a learned policy may select unsafe actions or exploit deficiencies in
the reward specification \cite{b16,b15}. Safety mechanisms can supplement
these policies. For instance, barrier-function shielding has been used for UAV
coverage with collision-avoidance and connectivity requirements
\cite{b38}, while predictive safety filters assess and modify proposed
inputs using a dynamical model \cite{b39}. In a shielded architecture,
the learned action is executed when it passes the model-based check;
otherwise, a backup controller supplies a replacement \cite{b16}.
Such interventions can restrict exploration, and their safety properties
depend on the information and model used for checking. 

Local motion planning alone does not determine how effectively a team
of robots executes their tasks to complete the \gls{sar} mission. Independent explorers may duplicate coverage, cluster around
the same frontier, or compete for narrow passages. Coordinated
exploration distributes robots across distinct targets \cite{b5},
while dynamic task allocation revises assignments as new tasks emerge
\cite{b6}. Event-driven supervisory control combines local decision
making with centralized intervention when mission conditions require
reallocation \cite{b7}. In \cite{b34}, centralized 
\gls{mpc} and local heuristic decision making are integrated to provide hierarchical coordination 
of individual robots for achieving both area coverage and victim localization under imperfect sensing.
This provides a precedent for connecting local planning with team
objectives.

Victim guidance introduces a further requirement, i.e., detected victims should be
willing and able to follow the robot \cite{b40, b41}. Trust influences reliance on
automation \cite{b8,b9}, while evacuation experiments show that people
may follow a robot despite previously observing unreliable behavior
\cite{b24}. The GT--BDI framework combines game theory with belief--desire--intention modeling to represent both 
individual decision making and social interaction of victims during evacuation, including the influence of trust on robot-assisted evacuation \cite{b10}. Complementary guidance research examines shepherding \cite{b11} and motion planning that balances exit-directed progress
with remaining visible to an evacuee \cite{b17}. These studies motivate evaluating guidance through evacuation outcomes alongside victim trust and following behavior.

\bigskip

These requirements are interdependent at the mission level:
conservative planning can restrict exploration, greater discovery
need not increase evacuation, and guidance depends on victim
responses. The resulting question is how improvements in local
control and team-level search translate into evacuation outcomes
under incomplete information.

We introduce \gls{pasar} as a mathematical and architectural framework for examining these mission-level relationships. Complementary simulation studies evaluate individual mechanisms and selected combinations.

\subsection*{Contributions}

The paper makes three contributions:

\begin{itemize}
\item
A mathematical formulation and architecture for mission-level
\gls{sar}:
We formulate \gls{pasar} as a perception-aware framework connecting
local motion planning, exploration, supervisory task allocation,
and adaptive approach-lead victim guidance. The formulation distinguishes environmental
ground-truth, robot-specific perceptions, and simulated victim
behavior, while separating component responsibilities from the
algorithms that implement them.

\item
An implementation of heterogeneous control and decision-making
mechanisms:
We implement model-predictive and learning-based exploration,
perception-based safety shielding, event-driven allocation, and adaptive
victim guidance within this architecture. This includes a
finite-action grid adaptation of state-dependent tube activation
and recurrent learned exploration with a model-predictive
safety backup.

\item An empirical investigation of local-control, coordination,
and evacuation trade-offs:
Complementary simulation studies distinguish the effects of
perception uncertainty, local constraint enforcement,
information sharing, task allocation, rescue-team size,
and adaptive guidance on search and evacuation outcomes.
\end{itemize}

The remainder of the paper is organized as follows.
Section~\ref{sec:problem-formulation} presents the architecture,
environment, robot and victim models, local controllers, and
supervisory coordination. Section~\ref{sec:experiments} reports the experimental
configurations and comparative results.
Section~\ref{sec:conclusion} summarizes the findings, limitations,
and directions for future research.

\section{PA-SaR: Perception-Aware Search-and-Rescue Architecture}
\label{sec:problem-formulation}

The Perception-Aware Search-and-Rescue (PA-SaR) architecture organizes
mission execution around a repeated observation--decision--execution
cycle. Each robot updates its local perception through automatic
environment scans, and its controller uses this information to
select the next control action. When supervisory coordination is
enabled, scan reports and mission events also update a shared
belief used by the supervisor to allocate exploration and rescue
tasks. Assigned targets and routes direct local planning;
unassigned robots explore autonomously. Once a rescue task enters
active guidance, a guidance policy replaces the robot's search
controller to guide the victim toward an exit.

Tables~\ref{tab:environmentvariables}, \ref{tab:robotvariables}, \ref{tab:victimvariables}, \ref{tab:mpcvariables}, \ref{tab:rlvariables}, \ref{tab:supervisoryvariables}, and \ref{tab:taskawarevariables} in this section define state, decision, and derived variables used in the mathematical formulation. Fixed model parameters, tuning coefficients, thresholds, and scenario-specific values are reported in Table~\ref{tab:model-variables}.

\subsection{Notation}

Throughout the paper the following rules are used for mathematical notations. Subscripts $r$ and $v$ identify robots and victims, respectively.
Where needed, upright superscripts $\mathrm r$ and $\mathrm v$
distinguish robot and victim variables. For a matrix $M$, $M_{ij}$
denotes its entry at row $i$ and column $j$. The notation
$\|\cdot\|_2$ denotes the Euclidean norm, and $|\mathcal S|$ denotes
the cardinality of a finite set $\mathcal S$. 
Predicted quantities carry the arguments $(k^{\prime}\mid k)$,
indicating a prediction for step $k^{\prime}\geq k$ based on the
information available at step $k$. 
The positive-part operator is $[z]_+=\max\{z,0\}$, and
$\operatorname{clip}(z,\ell,u)=\min\{u,\max\{\ell,z\}\}$
limits a scalar $z$ to the interval $[\ell,u]$, where $\ell\leq u$. Manhattan distance is the sum of the row and column separations.

\subsection{Modeling}

The following models define the ground-truth environment,
robot-specific perceptions, and robot and victim dynamics
within \gls{pasar}.

\subsubsection{Environment}
\label{subsec:environment}

\begin{table}[t]
    \centering
    \caption{Ground-truth environment notation.}
    \label{tab:environmentvariables}
    \begin{tabular*}{\columnwidth}{
        @{\extracolsep{\fill}}lp{0.63\columnwidth}@{}
    }
        \toprule
        \textbf{Variable} & \textbf{Description} \\
        \midrule
        $\nh$                         & Number of grid rows \\
        $\nw$                         & Number of grid columns \\
        $\mathcal{C}$                 & Set of grid cells \\
        $k$                           & Discrete step \\
        $\mathcal{R}$                 & Set of robots \\
        $\mathcal{V}(k)$              & Set of active victims \\
        $\mathcal{C}^{\mathrm{exit}}$ & Set of exit cells \\
        $\mtrav$                      & Traversability layer \\
        $\mijtrav$      & Traversability of cell $(i,j)$ \\
        $\mvuln$                      & Terrain-risk layer \\
        $\nusafe$         & Safe-terrain risk value \\
        $\numoderate$     & Moderate-terrain risk value \\
        $\nuhigh$         & High-risk-terrain value \\
        $\mvict(k)$                   & Victim occupancy layer \\
        $\mrobot(k)$                  & Robot occupancy layer \\
        $\mfire(k)$                   & Fire-state layer \\
        $\mburn(k)$                   & Burning-duration layer \\
        $\mijburn(k)$                 & Burning duration of cell $(i,j)$ \\
        $\bzallvict(k)$               & Collection of victim states \\
        $G(k)$                        & Ground-truth environment state \\
        $\mathcal{G}$                 & Set of admissible states \\
        \bottomrule
    \end{tabular*}
\end{table}

The ground-truth \gls{sar} environment is modeled in discrete space and
time. It occupies a predefined rectangular grid with $\nh$ rows and $\nw$
columns whose cell set
$\mathcal C
=
\left\{
[i\ j]^\top
 \middle|
i\in\{1,\ldots,\nh\},\
j\in\{1,\ldots,\nw\}
\right\}$
remains fixed throughout the simulation. Cell properties and occupancy evolve at discrete steps $k\in\mathbb N_0$.

The set of robots $\mathcal R$ is fixed, whereas $\mathcal V(k)$ denotes
the set of victims present in the environment at step $k$. This set is time dependent because it can change through evacuation or victim death, according to the victim model used. Robots and victims occupy cells in $\mathcal C$, with positions of
robot $r\in\mathcal R$ and victim $v\in\mathcal V(k)$ denoted by $\bq(k)=[i_r^{\mathrm{r}}(k)\ j_r^{\mathrm{r}}(k)]^\top$ and $\bqv(k) =[i^{\mathrm{v}}_v(k)\ j^{\mathrm{v}}_v(k)]^\top$, respectively. A fixed set $\mathcal C^{\mathrm{exit}}\subseteq\mathcal C$ contains the
exit cells through which victims can leave the environment.

The ground-truth environment is represented by co-registered
$\nh\times\nw$ spatial layers, each describing one cell-wise property: 
these include the traversability, vulnerability, victim, robot, and fire layers.

The traversability layer
$\mtrav\in\{0,1\}^{\nh\times\nw}$ represents fixed physical barriers.
An entry $\mijtrav=0$ denotes a traversable cell, whereas $\mijtrav=1$ denotes an obstructed cell, such as a wall
or blocked infrastructure.

The vulnerability layer
$\mvuln\in\{\nusafe, \numoderate, \nuhigh\}^{\nh\times\nw}$ assigns each cell a fixed, normalized terrain-risk level, where $\nusafe$, $\numoderate$, and $\nuhigh$ denote safe, moderately vulnerable, and high-risk terrain, respectively. 

Each entry of the binary victim and robot layers, $\mvict(k)$ and $\mrobot(k)$, denotes whether a cell 
is occupied by a victim or a robot. 
These entries are derived from the corresponding entity positions.

The fire layer $\mfire(k)\in \{0,1,2,3\}^{\nh\times\nw}$ distinguishes safe, flammable, burning, and burnt cells, represented by the categorical
values $0,1,2,3$ respectively. These categories and their transition rules follow a simplified,
discrete-time adaptation of the cellular-automaton
fire-propagation model in \cite{b35}. When a cell becomes burning, its traversable four-connected neighbors that are currently safe become flammable. At each subsequent fire-update
step, an ignition trial is performed with probability $\pignite$
for each adjacency between a burning cell and a flammable cell. A
flammable cell becomes burning if at least one corresponding ignition
trial succeeds. 
The burning duration matrix
$\mburn(k)\in\mathbb N_0^{\nh\times\nw}$ records the
number of fire-update steps elapsed since each currently burning cell ignited; its entries are zero for cells that are not burning. Once
$\mijburn(k)$ reaches the configured burning duration $\tburn$, cell $(i,j)$ transitions to the burnt state. Burnt cells neither ignite again nor propagate fire.

The ground-truth environment also includes the internal states of dynamic victims, defined in Section~\ref{subsec:victims}. These states retain the history required to determine
subsequent victim responses under the behavioral model.

The collection of spatial layers and internal variables required to model the ground-truth environment is

\begin{align}
\begin{split}
\label{eqn:groundstate}
    G(k)
    =
    \Big\{
        &\mtrav,
        \mvuln,
        \mfire(k),
        \mvict(k), \\
        &\mrobot(k),
        \mburn(k),
        \bzallvict(k)
    \Big\} \in \mathcal{G},
\end{split}
\end{align}
where $\mathcal G$ denotes the set of all admissible ground-truth realizations for set $G(k)$. 

Here, $\bzallvict(k)$ collects the individual states
$\bzvict_v(k)$ of all victims
$v\in\mathcal V(k)$, indexed by victim identity. This component
is omitted in stationary-victim configurations, where victim
locations are represented by the occupancy layer alone.
Because $G(k)$ is not directly available to the robots, each
robot maintains a partial, cell-wise perception of the
environment, introduced next.

\subsubsection{Robots}
\label{subsec:robot}

\begin{table}[t]
    \centering
    \caption{Robot perception and motion notation.}
    \label{tab:robotvariables}
    \begin{tabular*}{\columnwidth}{
        @{\extracolsep{\fill}}lp{0.63\columnwidth}@{}
    }
        \toprule
        \textbf{Variable} & \textbf{Description} \\
        \midrule
        $\bq(k)$
        & Robot position \\

        $\qrobot(k)$
        & Robot perception state \\

        $\qsetrobot$
        & Set of admissible perception states \\

        $\ptrav(k)$
        & Perceived traversability layer \\

        $\pvuln(k)$
        & Perceived vulnerability layer \\

        $\pvict(k)$
        & Perceived victim layer \\

        $\probot(k)$
        & Perceived robot layer \\

        $\pfire(k)$
        & Perceived fire layer \\

        $\pconf(k)$
        & Confidence layer \\

        $\pexpl(k)$
        & Exploration layer \\

        $\varepsilon_{r,ij}$
        & Vulnerability measurement error \\

        $\boldsymbol{c}_{ij}$
        & Position of cell $(i,j)$ \\

        $\pijconf(k)$
        & Confidence in cell $(i,j)$ \\

        $\eta_{r,ij}(k)$
        & Scan quality for cell $(i,j)$ \\

        $\bu(k)$
        & Commanded control input \\

        $\mathcal{U}$
        & Set of control inputs \\

        $(i^{\prime},j^{\prime})$
        & Post-command cell \\

        $\ptrip(k)$
        & Displacement probability \\

        $\mijprimevuln$
        & Vulnerability of cell $(i^{\prime},j^{\prime})$ \\

        $\pijprimeconf(k)$
        & Confidence in cell $(i^{\prime},j^{\prime})$ \\
        \bottomrule
    \end{tabular*}
\end{table}

Robots may switch between autonomous exploration (\emph{idle}),
assigned exploration (\emph{explore}), and victim rescue
(\emph{rescue}), subject to their capabilities and the task-transition rules described in
Section~\ref{subsec:supervisorycoordination}. A cell is considered \emph{observed} by robot $r\in\mathcal R$ if it lies within the robot's sensing radius $\rrobot$ and the line of sight from the robot is not blocked by an intervening obstructed cell. 

At all steps, each robot is assumed to know its own position $\bq(k)\in\mathcal C$ exactly. Its
stored information about the environment at step $k$ is represented by
the perception state given by 

\begin{align}
\begin{split}
\label{eqn:robot-perception-state}
\qrobot(k)
=
\bigl\{
&\ptrav(k),
\pvuln(k), \\
&\pvict(k),
\probot(k),
\pfire(k), \\
&\pconf(k), 
\pexpl(k)
\bigr\}
\in \qsetrobot,
\end{split}
\end{align}
where $\qsetrobot$ denotes the set of admissible robot-perception states. 
The first five components are perceived versions of the corresponding
ground-truth spatial layers in Section~\ref{subsec:environment},
while $\pconf(k)$ and $\pexpl(k)$ store cell-wise confidence and
exploration status, respectively. All matrices in $\qrobot(k)$ have dimensions $\nh\times\nw$, and the sensing radius $\rrobot$ limits which cells can be observed.

Before the initial scan at $k=0$, all perceived environment layer entries
are initialized to zero except for the robot's self-occupancy marker.
The confidence and exploration layers are also initialized to zero.
Thus, unobserved cells have an optimistic categorical prior, but are
marked as unsupported by observations.

For observed cells, traversability, victim occupancy, robot
occupancy, and fire are measured without error and replace the
corresponding stored values. This assumption does not imply
complete environment knowledge: cells outside the observed region
retain their previous values, which may be outdated.

Vulnerability measurements include additive noise. The value
stored by robot $r$ for observed cell $(i,j)$ is
$\operatorname{clip}(\mijvuln+\varepsilon_{r,ij},0,1)$, where
$\varepsilon_{r,ij}\sim
\mathrm{Unif}[-\deltavuln,
\deltavuln]$
and $\deltavuln\geq0$ is the measurement-noise bound.
The self-occupancy marker is updated after every realized movement.

For each cell, the exploration entry is set to one upon its first
observation and remains one thereafter. A single confidence value,
shared by the perceived environment layers at that cell, heuristically
represents information quality and freshness. It is updated independently without modifying the stored measurements.

Let $\boldsymbol c_{ij}=[i\ j]^\top$ denote the position of cell $(i,j)$. The confidence value corresponding to robot $r$ and associated with cell $(i,j)$ evolves according to \eqref{eqn:pconfijupdate}.
\begin{figure*}
\begin{equation}
\small
\label{eqn:pconfijupdate}
\pijconf(k+1) =
\begin{cases}
\pijconf(k) + \eta_{r,ij}(k+1) \left(
1-\pijconf(k)
\right), & \text{cell $(i,j)$ observed at $k+1$} \\
\max\left\{
0,
\pijconf(k)-\sigmaaconf
\right\},
& \text{otherwise}.
\end{cases}
\end{equation}
\end{figure*}
In \eqref{eqn:pconfijupdate}, $\sigmaaconf\geq0$ is the confidence-decay parameter and
$\eta_{r,ij}(k+1)$ is the scan-quality factor inspired by \cite{b34}. When distance falloff
is enabled, this factor is

\begin{equation}
\label{eqn:scanquality}
\eta_{r,ij}(k+1)
=
\eta_r^0
\exp\left(
-\frac{
\norm{\boldsymbol c_{ij}-\bq(k+1)}^2_2
}{
2(\rrobot/3)^2
}
\right),
\end{equation}
where $\eta_r^0\in[0,1]$ is the base scan quality. When distance
falloff is disabled, $\eta_{r,ij}(k+1)=\eta_r^0$ for every observed
cell. 

\bigskip

After the scan, the controller associated with robot $r$ selects a commanded control input $\bu(k)\in\mathcal U$, where

\begin{equation}
\label{eqn:controlinputset}
\mathcal U
=
\left\{
\begin{bmatrix}-1\\0\end{bmatrix},
\begin{bmatrix}1\\0\end{bmatrix},
\begin{bmatrix}0\\-1\end{bmatrix},
\begin{bmatrix}0\\1\end{bmatrix},
\begin{bmatrix}0\\0\end{bmatrix}
\right\},
\end{equation}
corresponding to moving \texttt{up}, \texttt{down}, \texttt{left},
\texttt{right}, and \texttt{wait}, respectively. Whether feasibility is considered during command selection depends on the
controller type and is elaborated in Section~\ref{subsec:controllers}. The robot subsequently attempts to execute the selected input.

After command processing, the robot may experience an uncontrolled displacement by one
cardinal cell. The probability of such a displacement occurring increases with ground-truth vulnerability and
decreases with confidence, which is assumed to represent
awareness that reduces susceptibility to the hazard. Let $(i^{\prime},j^{\prime})$ index the cell occupied by robot $r$ after command processing and before any displacement. The displacement probability is evaluated at this cell:

\begin{align}
\begin{split}
\label{eqn:slip-chance}
&\ptrip(k) \\
&=
\left[
\mijprimevuln
-
\alphatrip
\pijprimeconf(k)
\right]_+^{\betatrip},
\end{split}
\end{align}

where $\alphatrip\geq0$ controls the influence of confidence and
$\betatrip>0$ shapes the dependence on the remaining risk.
Because vulnerability and confidence lie in $[0,1]$, the
positive-part operator ensures that the bracketed quantity lies
in $[0,1]$, and hence $\ptrip(k)\in[0,1]$. The parameter values
are specified in the experimental configurations.

Conditional on a motion disturbance occurring, a one-cell
displacement direction is sampled uniformly from the four
cardinal directions. This displacement is imposed independently
of the controller's command, unless blocked by the execution
constraints described in
Section~\ref{subsec:common-experimental-setup}.

\subsubsection{Victims}
\label{subsec:victims}

\begin{table}[t]
    \centering
    \caption{Victim state and behavior notation.}
    \label{tab:victimvariables}
    \begin{tabular*}{\columnwidth}{
        @{\extracolsep{\fill}}lp{0.63\columnwidth}@{}
    }
        \toprule
        \textbf{Variable} & \textbf{Description} \\
        \midrule
        $\zvict(k)$
        & Victim state \\

        $\bqv(k)$
        & Victim position \\

        $\kappa_v$
        & Victim age group \\

        $\pi_v(k)$
        & Interaction strategy \\

        $\theta_v(k)$
        & Victim trust \\

        $\delta_v^{\mathrm{trust}}$
        & Trust sensitivity \\

        $\qvictim(k)$
        & Victim perception state \\

        $h_v(k)$
        & Victim history \\

        $S_v(k)$
        & Perceived safety rank \\

        $\Delta S_v(k)$
        & Change in perceived safety \\

        $d_v(k)$
        & Distance to nearest perceived robot providing active rescue guidance. \\

        $\Delta d_v^+(k)$
        & Robot approach signal \\

        $\theta_v^{\mathrm{base}}(k)$
        & Baseline trust \\

        $w_v^{\mathrm{safe}}(k)$
        & Applied safety-change weight \\
        \bottomrule
    \end{tabular*}
\end{table}

Victims may be stationary or dynamic. A stationary victim remains at its
initial position, whereas a dynamic victim evaluates at each simulation step whether and
where to move. The remainder of this subsection describes the
dynamic-victim model, which is adapted from the GT--BDI framework
\cite{b10}. This framework combines individual belief-based decision
making with a game-theoretic mechanism for resolving movement conflicts
between victims.

For each dynamic victim $v\in\mathcal V(k)$, the individual state
$\zvict(k)$ comprises its position $\bqv(k)$, age group $\kappa_v \in \left\{
\mathrm{child},
\mathrm{adult},
\mathrm{elderly}
\right\}$,
%personality 
interaction strategy $\pi_v(k) \in \left\{
\mathrm{altruistic},
\mathrm{egoistic},
\mathrm{selfish}
\right\}$, trust state $\theta_v(k)$, trust sensitivity $\delta_v^{\mathrm{trust}}$,
perceived environment $\qvictim(k)$, and %behavioral 
history $h_v(k)$.
The age group is fixed, while $\delta_v^{\mathrm{trust}}$ is determined by the initial
interaction strategy $\pi_v(0)$ and remains unchanged if the  interaction strategy later
changes (Table~\ref{tab:model-variables}). The history stores preceding perceived safety, the distance to and position of the nearest perceived rescue robot, the victim's preceding position, and counters recording consecutive stationary steps for the trust-reset rules below.

The current interaction strategy determines autonomous movement preferences and conflict priorities
(Table~\ref{tab:personality}). The trust state $\theta_v(k)\in[0,1]$ represents the victim's willingness to follow
a perceived robot and is not associated with a particular robot. The sensitivity $\delta_v^{\mathrm{trust}}>0$ scales the
trust response to changes in robot proximity and perceived safety.

\begin{table*}[t]
\centering
\caption{Victim interaction strategy types and associated behavior.}
\label{tab:personality}

\begin{tabularx}{\textwidth}{@{}lYY@{}}
\toprule
\textbf{Interaction strategy $\pi_v$}
&
\textbf{Autonomous preference when no danger is perceived}
&
\textbf{Conflict-resolution strategy}
\\
\midrule

Altruistic
&
Approaches nearby victims
&
Yields to egoistic and selfish victims
\\

Egoistic
&
Approaches nearby victims
&
Yields only to selfish victims
\\

Selfish
&
Moves toward the exit
&
Takes priority over victims using other interaction strategies
\\

\bottomrule
\end{tabularx}
\end{table*}

Victims use the same radius- and line-of-sight-based scanning rule as the
robots, with sensing radius $\rvict$. Their perceived environment $\qvictim(k)$ contains
traversability, vulnerability, fire, victim occupancy, and robot
occupancy. Victim decisions
use these observations directly, without confidence weighting or an exploration objective.

Using its perceived environment $\qvictim(k)$,  victim $v$ classifies the
perceived safety of its current cell and four-connected neighboring
cells. Traversable cells are ranked, in increasing order of safety, as
burning or burnt, high-risk, flammable, moderately vulnerable, and
hazard-free, with the least-safe applicable classification taking
precedence. The victim perceives danger when its current cell is burning, burnt, high-risk, or flammable.

For use in the trust update model, these safety classifications are mapped to
consecutive ordinal ranks in the same order. Let $S_v(k)$ denote the rank assigned at step $k$ to the victim's current cell. The
change in perceived safety is
$\Delta S_v(k)=S_v(k)-S_v(k-1)$, where $S_v(k-1)$ is the rank assigned at
step $k-1$ to the cell occupied by the victim at that step.

Trust and following decisions refer to the nearest robot perceived as providing active rescue guidance. When such a robot is perceived, $d_v(k)$ denotes its Euclidean distance from the victim. The approach signal $\Delta d_v^+(k)=[d_v(k-1)-d_v(k)]_+$
is set to zero if no such robot was perceived at the preceding step. 

The baseline $\theta_v^{\mathrm{base}}(k)$ is the trust value
to which the current proximity and safety contributions are
added. It normally equals $\theta_v(k-1)$.
Let $\theta^{\mathrm{follow}}\in[0,1]$ denote the trust
threshold for following a perceived rescue robot. If the perceived rescue robot remains stationary for its
prescribed timeout, the baseline is reset to zero. Otherwise,
if the victim remains stationary for its prescribed timeout,
has preceding trust
$\theta_v(k-1)\geq\theta^{\mathrm{follow}}$,
and the robot remains within sensing range, the baseline is
reset to $\theta^{\mathrm{follow}}/2$.
Timeout values are given in Table~\ref{tab:model-variables}. 

When a rescue robot is perceived, the resulting trust update is

\begin{align}
\begin{split}
\label{eqn:victim-trust-update}
\theta_v(k)
&=
\operatorname{clip}
\Big(
\theta_v^{\mathrm{base}}(k)
+
\delta_v^{\mathrm{trust}} \\
&\left[
\wprox \Delta d_v^+(k)
+
w_v^{\mathrm{safe}}(k)\Delta S_v(k)
\right],
0,1
\Big).
\end{split}
\end{align}

Here, $\wprox$ weights the approach signal and 
$w_v^{\mathrm{safe}}(k)=w^{\mathrm{gain}}$ for
$\Delta S_v(k)\geq0$ and
$w_v^{\mathrm{safe}}(k)=w^{\mathrm{loss}}$ otherwise. The weights satisfy
$w^{\mathrm{loss}}>w^{\mathrm{gain}}\geq0$,
giving worsening safety greater influence than an equally
sized improvement \cite{b23}. If no robot providing active rescue guidance is perceived,
trust remains unchanged, while the stored safety history is refreshed. 

At each step, a victim either moves to a cardinally adjacent
cell or remains in its current cell. For personal-space
responses and movement toward a robot or another victim,
a neighboring cell is \emph{perceptually admissible} if it
lies inside the grid, is perceived as traversable, and is
neither burning nor burnt.

After updating trust, a victim that perceives a robot in a
cardinally adjacent cell selects an admissible neighbor
maximizing its Manhattan distance from that robot, if one
exists. This personal-space movement takes precedence over
the ordinary movement trial, regardless of trust.

If no personal-space movement is selected, an age-dependent Bernoulli trial determines whether the victim attempts an ordinary movement. Age-dependent walking speeds
\cite{b20,b21} and observed changes in movement under emergency
conditions \cite{b22} motivate this simplified rule. In the model,
the movement probability $p^{\mathrm{move,}\kappa_v}$ is highest for
adults, followed by elderly victims and children, and is multiplied
by $\mu^{\mathrm{danger}}$ when danger is perceived, with the resulting probability capped
at one. These probabilities
and their ordering are modeling choices. If the trial fails, the victim does not move.

If the trial succeeds, the victim follows the nearest perceived
rescue robot when
$\theta_v(k)\geq\theta^{\mathrm{follow}}$
and an admissible neighbor exists. It selects the neighbor
minimizing its Manhattan distance to that robot. Otherwise, it uses the autonomous movement rules below.

During autonomous movement, the interaction-strategy-dependent preference in
Table~\ref{tab:personality} applies unless danger is perceived, in which case every victim prioritizes the nearest exit.
Exit-directed movement first considers moderately vulnerable
or hazard-free cells that reduce distance to the exit,
followed by detours within these safety categories,
non-burning exit-directed cells, and finally any
perceived-traversable neighbor. The victim waits if none exists.

For victim-directed movement, the target is the nearest perceived victim. If that victim lies in one of the eight surrounding cells, the exit-directed rule applies instead. If no victim is perceived or no
perceptually admissible victim-directed movement exists, the victim randomly selects a moderately vulnerable or hazard-free neighbor, or waits if none exists. 

A selected victim movement is canceled if its destination
is occupied by a robot or another victim at the beginning
of the step. If multiple victims select the same available
destination, only the highest-priority victim moves there,
with uniform random tie-breaking among victims of equal
highest priority. The others remain in their current cells
for that step, without selecting alternative destinations.
An egoistic victim competing with a selfish victim becomes
selfish for subsequent decisions.

\subsection{Controllers}
\label{subsec:controllers}

This subsection defines the controllers that generate commanded robot
movements from their current position $\bq(k)$ and perception state $Q_r^{\mathrm{r}}(k)$. The controllers comprise three model-predictive variants, a learned
policy, and a hybrid shielded policy. 

\subsubsection{Model-predictive controllers}
\label{subsubsec:modelpredictivecontrollers}

\begin{table}[t]
    \centering
    \caption{\gls{mpc} prediction and objective notation.}
    \label{tab:mpcvariables}
    \begin{tabular*}{\columnwidth}{
        @{\extracolsep{\fill}}lp{0.57\columnwidth}@{}
    }
        \toprule
        \textbf{Variable} & \textbf{Description} \\
        \midrule
        $k^{\prime}$
        & Predicted step \\

        $\bq(k^{\prime}\mid k)$
        & Predicted robot position \\

        $\boldsymbol{U}_r(k)$
        & Candidate control sequence \\

        $\boldsymbol{u}_r(k^{\prime}\mid k)$
        & Predicted control input \\

        $\mathcal{U}^N$
        & Set of length-$N$ input sequences \\

        $J_r(\boldsymbol{U}_r(k))$
        & Sequence objective value \\

        $L_r(k^{\prime}\mid k)$
        & Stage objective \\

        $V_r^{\mathrm{terminal}}(k+N\mid k)$
        & Terminal objective \\

        $\phidisc(k^{\prime}\mid k)$
        & Discovery score \\

        $\phiconf(k^{\prime}\mid k)$
        & Confidence score \\

        $\phivuln(k^{\prime}\mid k)$
        & Vulnerability score \\

        $\mathcal{O}_r(k^{\prime}\mid k)$
        & Cells predicted to be observed \\

        $\mathcal{E}_r(k^{\prime}-1\mid k)$
        & Previously explored cells \\

        $\zdiscover$
        & Discovery normalization factor \\

        $\pijconf(k^{\prime}\mid k)$
        & Predicted cell confidence \\

        $\pijvuln(k^{\prime}\mid k)$
        & Predicted cell vulnerability \\

        $d_r^{\mathrm{front}}$
        & Normalized frontier distance \\

        $\pijexpl(k)$
        & Cell exploration indicator \\
        \bottomrule
    \end{tabular*}
\end{table}

The three model-predictive controllers use the same receding-horizon
procedure. At step $k$, each controller evaluates sequences of $N$
candidate inputs. Candidate positions are obtained by applying the commanded
movements along the prediction horizon. The displacement is not added
to this nominal trajectory; instead, its possible effects are represented
by the controller-specific feasibility conditions described below. Only
the first input of the selected sequence is applied, after which the
robot scans again and the optimization is repeated.

Perceived traversability and victim occupancy are held
constant throughout the prediction horizon; victim motion
is not predicted. Perceived vulnerability also remains
unchanged unless measurement noise is enabled, in which
case predicted scans perturb the stored vulnerability
values of cells predicted to be observed.

Candidate movements update the robot's predicted position and self-occupancy. At each predicted position, the sensing-radius
and line-of-sight rules discussed in Section~\ref{subsec:robot} determine which cells would be observed, using perceived traversability to determine occlusion.
Confidence is then updated according to
\eqref{eqn:pconfijupdate}, with scan quality evaluated from
the predicted position using \eqref{eqn:scanquality} when
distance falloff is enabled. Predicted observations mark
cells as explored, while previously explored cells remain
explored. These updates model information acquisition without
accessing previously unknown ground-truth cell values.

Fire propagation is predicted using an internal stochastic forecast with assumed ignition probability $\pignitehat$, which may differ from the ground-truth value $\pignite$.

When burnout prediction is enabled, the controller additionally maintains a local counter for each cell perceived as burning. The counter increases while the cell remains unobserved and is reset upon observation. A burning cell is predicted to become burnt when its counter reaches the assumed threshold $\tburnhat$.

These counters are not advanced along the prediction horizon, so elapsed prediction time alone does not induce burnout. Without this extension, the forecast predicts fire spread without burnout. The prediction settings used in each study are specified in Table~\ref{tab:model-variables}.

Within the prediction model, a grid cell is classified as feasible if it is perceived as traversable and free of victims and is not predicted to be burning. Perceived robot occupancy is not included in this common feasibility condition because the controllers do not jointly predict the movements of the other robots. 

Nominal \gls{mpc} requires only the predicted nominal positions to remain
feasible. It can therefore select a movement whose commanded destination
is feasible even when a displacement from that destination could reach
an infeasible cell. This controller provides the reference against which
the additional robustness and conservativeness of the tube-based
controllers are evaluated.

% \paragraph{Fixed-tube TMPC.}

Fixed-tube \gls{tmpc} requires both the commanded destination and every
cell that can be reached from it through a one-cell cardinal displacement to be
feasible. The same one-cell tube is imposed at every predicted position,
irrespective of the local vulnerability or confidence. This provides
worst-case local protection against the modeled displacement disturbance but may
reject narrow passages in which the nominal trajectory itself is
feasible.

% \paragraph{State-dependent dynamic-tube MPC.}

The \gls{sddtmpc} controller uses perceived vulnerability and confidence
to determine whether a displacement should be considered along a candidate
trajectory. At prediction step $k^{\prime}$, let
$\bq(k^{\prime}\mid k)=\boldsymbol c_{ij}$ denote the candidate
position for robot $r$ before accounting for a possible stochastic displacement. The controller estimates the displacement probability using
\eqref{eqn:slip-chance}, replacing ground-truth vulnerability
and current confidence with their predicted perceived values
$\pijvuln(k^{\prime}\mid k)$ and
$\pijconf(k^{\prime}\mid k)$.
It imposes the one-cell tube whenever this estimate is positive. The tube size is therefore binary in the present grid model. This is a discrete-space adaptation of the continuously
varying \gls{sddtmpc} formulation in \cite{b4}: fractional tube radii
have no direct interpretation when all disturbances are whole-cell
displacements.

The resulting finite-horizon planning problem is a discrete,
non-convex combinatorial optimization problem. Comparable
search-and-rescue path-planning problems are commonly solved using
numerical search methods \cite{b34}. 
The finite action set permits exhaustive enumeration of the candidate
sequences; infeasible branches are discarded and the feasible sequence
with the greatest objective value is selected.

For a prediction horizon $N$, let
$\boldsymbol U_r(k)=
\bigl(
\boldsymbol u_r(k\mid k),\ldots,
\boldsymbol u_r(k+N-1\mid k)
\bigr)\in\mathcal U^N$
denote a candidate control sequence for robot $r$. Its objective value is
\begin{align}
\begin{split}
\label{eqn:mpc-objective}
J_r \left(\boldsymbol U_r(k)\right)
=
\sum_{k^{\prime}=k+1}^{k+N}
&\left(\gamma^{\mathrm{MPC}}\right)^{k^{\prime}-k-1}L_r(k^{\prime}\mid k) \\
&+
V_r^{\mathrm{terminal}}(k+N\mid k).
\end{split}
\end{align}
The discount factor $\gamma^{\mathrm{MPC}} \in(0,1]$ determines the relative importance of predictions farther into the horizon, $L_r(\cdot)$ represents the stage objective, also called the exploration objective, and the terminal term $V_r^{\mathrm{terminal}}$ adds a contribution at the final predicted state.

The three model-predictive controllers use a common exploration
objective that rewards newly observed cells and higher predicted
confidence values while including a vulnerability term. The stage objective is given by

\begin{align}
\begin{split}
\label{eqn:mpc-stage-objective}
L_r(k^{\prime}\mid k)
&=
\wdisc \phidisc(k^{\prime}\mid k) \\
&+
\wconf \phiconf(k^{\prime}\mid k) \\
&-
\wvuln \phivuln(k^{\prime}\mid k).
\end{split}
\end{align}

The non-negative weights $\wdisc$, $\wconf$, and $\wvuln$ balance predicted discovery, confidence, and vulnerability scores. Table~\ref{tab:mpc-objective-components} defines these scores.
The discovery score $\phidisc$ counts cells predicted to be observed
for the first time along the candidate trajectory. The
confidence score $\phiconf$ averages the predicted confidence layer
obtained from \eqref{eqn:pconfijupdate}. The vulnerability
score $\phivuln$ is the predicted perceived vulnerability at the
candidate robot position. Scaling is included in the score definitions; therefore, weights specify relative importance.

\begin{table*}[t]
\centering
\caption[Discovery, confidence, and vulnerability components of the MPC stage objective.]{
Discovery, confidence, and vulnerability components of the
\gls{mpc} stage objective.
The sets $\mathcal O_r(k^{\prime}\mid k)$ and
$\mathcal E_r(k^{\prime}-1\mid k)$ contain the cells predicted to be
observed at step $k^{\prime}$ and explored before that step, respectively.
The discovery scale $\zdiscover=\pi(\rrobot)^2/2$ equals
half the area of a circle with radius $\rrobot$. It provides a geometric reference scale.
}
\label{tab:mpc-objective-components}

\footnotesize
\begingroup
\setlength{\tabcolsep}{3pt}
\renewcommand{\arraystretch}{1.45}

\begin{tabularx}{\textwidth}{
    @{}
    >{\raggedright\arraybackslash}p{0.20\textwidth}
    >{\raggedright\arraybackslash}X
    @{}
}
\toprule
\textbf{Component} & \textbf{Definition} \\
\midrule

Discovery
&
\(
\begin{aligned}
\phidisc(k^{\prime}\mid k)
&=
\frac{
    1
}{
    \zdiscover
} \left|
        \mathcal O_r(k^{\prime}\mid k)
        \setminus
        \mathcal E_r(k^{\prime}-1\mid k)
    \right|
\end{aligned}
\)
\\

Confidence
&
\(
\begin{aligned}
\phiconf(k^{\prime}\mid k)
&=
\frac{1
}{
    \nh\nw
} \sum_{i=1}^{\nh}
    \sum_{j=1}^{\nw}
    \pijconf(k^{\prime}\mid k)
\end{aligned}
\)
\\

Vulnerability
&
\(
\begin{aligned}
\phivuln(k^{\prime}\mid k)
&=
\pijvuln(k^{\prime}\mid k),
&
\bq(k^{\prime}\mid k)
&=
\boldsymbol c_{ij}
\end{aligned}
\)
\\

\bottomrule
\end{tabularx}

\endgroup
\end{table*}

The terminal contribution for robot $r$ is
\begin{align}
\small
\begin{split}
\label{eqn:mpc-terminal-objective}
&V_r^{\mathrm{terminal}}(k+N\mid k) \\
&=
-\wterm N
d_r^{\mathrm{front}}
\left(
\bq(k+N\mid k);Q_r^{\mathrm{r}}(k)
\right).
\end{split}
\end{align}

Here, $\wterm\geq0$ is the terminal-objective weight.
The quantity $d_r^{\mathrm{front}}$ is the four-connected
shortest-path distance from the terminal predicted position to a cell marked as unexplored in robot $r$'s exploration layer, i.e. $\pijexpl(k)=0$, normalized by $\nh+\nw$.

This distance is computed using the perception available
at step $k$. The path may traverse only
cells satisfying the controller-specific feasibility
conditions. If no unexplored cell is reachable, an implementation-specific fallback
value is used, as specified in Table~\ref{tab:model-variables}.

Constraint satisfaction is assessed with respect to the
robot's perception and prediction model. Occlusion,
measurement noise, outdated information, and fire-prediction
errors can therefore cause discrepancies between predicted
feasibility and actual conditions.

Recursive feasibility would require a feasible planning problem
to remain feasible at subsequent decision steps under the
modeled evolution. No terminal invariant set or alternative
recoverability mechanism is imposed here, and no recursive-
feasibility proof is provided. The formulation therefore
establishes neither an unconditional ground-truth safety
guarantee nor a formal recursive-feasibility guarantee.

\glsreset{rl}
\subsubsection{\gls{rl} controller}
\label{subsubsec:rl-controller}

\begin{table}[t]
    \centering
    \caption{\gls{rl} reward notation.}
    \label{tab:rlvariables}
    \begin{tabular*}{\columnwidth}{
        @{\extracolsep{\fill}}lp{0.57\columnwidth}@{}
    }
        \toprule
        \textbf{Variable} & \textbf{Description} \\
        \midrule
        $R_r(k)$
        & Transition reward \\

        $\Delta N^{\mathrm{detected}}(k)$
        & Number of newly detected victims \\

        $\Delta C_r^{\mathrm{terrain}}(k)$
        & Terrain-collision increment \\

        $\Delta C_r^{\mathrm{victim}}(k)$
        & Victim-collision increment \\

        $E_r(k)$
        & Explored traversable-cell fraction \\

        $\mathbf{1}^{\mathrm{success}}(k)$
        & Successful-completion indicator \\

        $\mathbf{1}^{\mathrm{end}}(k)$
        & Episode-termination indicator \\

        $\Omega_r(k)$
        & Coverage potential \\
        \bottomrule
    \end{tabular*}
\end{table}

The \gls{rl} controller provides a learned alternative to
model-predictive action selection and can be combined with a
model-based safety shield. The task is partially observable:
the policy receives the robot's perceived environment and known
position, but not the ground-truth environment or the current
time-step index.

The policy observation is an 11-channel spatial encoding of the
perception state: one channel each for perceived traversability,
vulnerability, victim occupancy, and robot occupancy; four one-hot
channels for the perceived fire states; the confidence and exploration
layers; and a normalized distance field to the nearest perceived
unexplored region. The policy selects an action from \eqref{eqn:controlinputset}. A recurrent policy retains information from previous observations, motivated by recurrent approaches to partially observable decision problems \cite{b29}. Its actor and critic are trained using
\gls{ppo} \cite{b28}, with generalized advantage estimation
\cite{b33}.

The reward promotes victim detection, efficient exploration, and timely
completion while discouraging collisions. Let
$\Delta N^{\mathrm{detected}}(k)$ denote the team-wide number of previously
undetected victims first detected by at least one robot during the
transition from $k$ to $k+1$. A victim detected by multiple robots is
counted only once. Furthermore, $\Delta C_r^{\mathrm{terrain}}(k)$ and $\Delta C_r^{\mathrm{victim}}(k)$ denote the increments in 
the cumulative terrain-collision and victim-collision counts for robot $r$ during the same transition. Finally, $E_r(k)\in[0,1]$ is the fraction
of ground-truth traversable cells that robot $r$ has marked as explored by step $k$. Then, the transition
reward is defined as
\begin{align}
\begin{split}
\label{eqn:rl-reward}
R_r(k)
={}&
\wdet
\Delta N^{\mathrm{detected}}(k)
-
\wstep
\\ &-
\wterr
\Delta C_r^{\mathrm{terrain}}(k)
-
\wvict
\Delta C_r^{\mathrm{victim}}(k) \\
&+
\wsucc
\mathbf 1^{\mathrm{success}}(k) \\
&+
\wcov
E_r(k+1)
\mathbf 1^{\mathrm{end}}(k) \\
&+
\gamma^{\mathrm{RL}} \Omega_r(k+1)
-
\Omega_r(k).
\end{split}
\end{align}

All reward weights are non-negative and set the relative importance
and scale of the terms indicated by their superscripts. They do not necessarily sum to one. Coverage is already expressed as a fraction,
whereas detection and collision increments are counts.
The coefficient $w^\Omega\geq0$ scales the coverage potential,
and $\gamma^{\mathrm{RL}}\in(0,1]$ is the discount factor used
for policy training and potential-based shaping.

The indicator $\mathbf 1^{\mathrm{success}}(k)$ equals one
on the transition that completes detection of all victims and
terminates the episode; it is zero on earlier transitions and
when the episode ends without success.
The indicator $\mathbf 1^{\mathrm{end}}(k)$ equals one on the
final transition, whether termination results from successful
completion or expiration of the episode horizon.

For nonterminal states, we define the coverage potential as
$\Omega_r(k)=-w^\Omega\bigl(1-E_r(k)\bigr)$ and assign zero
potential to terminal states. The terminal-coverage term rewards
the coverage achieved when the episode ends. The shaping term
$\gamma^{\mathrm{RL}}\Omega_r(k+1)-\Omega_r(k)$ follows the
potential-based reward-shaping formulation of \cite{b30},
with coverage chosen here as the basis of the potential.

Ground-truth traversability is used to calculate coverage during
training, but is not provided to the policy or safety shield.
The unshielded configuration issues the policy's proposed command
unchanged. The shielded configuration checks the proposal against
the robot's current perceived environment and, if it fails,
invokes the fixed-tube \gls{tmpc} controller as a backup.

The proposed destination must lie inside the grid, be perceived
as traversable and victim-free, not be perceived as burning,
and have perceived vulnerability below
$\theta^{\mathrm{shield}}$.
These conditions use the information available when the command
is selected; the shield does not predict fire spread before
arrival. The victim, fire, and vulnerability conditions also
apply to every in-grid, four-connected neighbor of the destination
that is perceived as traversable, accounting for possible
one-cell disturbances. Out-of-grid and perceived-obstructed
neighbors are excluded because displacements into such cells
are blocked under the motion model.

The fixed-tube \gls{tmpc} backup is evaluated only when the shield
rejects a proposal. It enforces the grid, traversability, victim,
and fire constraints over the complete one-cell tube, but treats
vulnerability as an objective term rather than a hard constraint.
Its command may therefore exceed the shield's vulnerability
threshold; if no feasible backup sequence exists, the robot waits.
Neither the shield nor the backup provides an unconditional
ground-truth safety guarantee, because both rely on perception.
The issued command remains subject to the movement-execution
rules described in
Section~\ref{subsec:common-experimental-setup}.

\subsubsection{Supervisory coordination}
\label{subsec:supervisorycoordination}

\begin{table}[t]
    \centering
    \caption{Supervisory task-allocation notation.}
    \label{tab:supervisoryvariables}
    \begin{tabular*}{\columnwidth}{
        @{\extracolsep{\fill}}lp{0.56\columnwidth}@{}
    }
        \toprule
        \textbf{Variable} & \textbf{Description} \\
        \midrule
         $\tau$
        & Candidate task type \\
        
        $\tasktype(k)$
        & Robot task type \\

        $\taskstatus(k)$
        & Rescue execution status \\

        $n$
        & Frontier-component size \\

        $m$
        & Number of frontier targets \\

        $\mathcal R_i^{\mathrm{available}}(k)$
        & Available robots at iteration $i$ \\

        $\mathcal T_i^{\mathrm{available}}(k)$
        & Available tasks at iteration $i$ \\

        $\bqtask(k)$
        & Task target position \\

        $U_{r,t,i}^\tau(k)$
        & Assignment utility \\

        $S_{t,i}(k)$
        & Spatial-redundancy penalty \\

        $\mathcal R_i^{\mathrm{idle}}(k)$
        & Unavailable idle robots \\

        $\mathcal T_i^{\mathrm{explore}}(k)$
        & Assigned exploration tasks \\

        $\bqtaskprime(k)$
        & Assigned exploration target \\

        $\bqprime(k)$
        & Last reported robot position \\

        $\Pi_r(k)$
        & Assigned path \\

        $\boldsymbol q_{r,j}^{\mathrm{path}}$
        & Path cell with index $j$ \\
        \bottomrule
    \end{tabular*}
\end{table}

The centralized supervisor coordinates exploration and rescue
through a shared environment belief and robot and victim records. Robots communicate scan results and mission events to the supervisor rather than communicating directly. These reports support task generation and allocation, while event-triggered intervention determines when assignments are reconsidered.

Each task $t$ has a type
$\tau\in\{\mathrm{explore},\mathrm{rescue}\}$.
A robot without an assignment is idle and explores autonomously. For an assigned robot $r$, $\tasktype(k)$ denotes its current task type. During rescue,
$\taskstatus(k)\in\{\mathrm{searching},
\mathrm{active\text{-}guidance}\}$ distinguishes travel toward
the victim from guidance toward an exit.

\noindent
\textbf{Shared belief and mission records:} The supervisor maintains a shared perception containing traversability,
vulnerability, victim occupancy, robot occupancy, fire, confidence, and
exploration layers analogous to the robot perceptions introduced in Section~\ref{subsec:robot}. Before receiving the first scans, all entries are
initialized to zero.

After each transition, the supervisor processes the robot scans
sequentially. For each reported cell, the last processed report
overwrites the stored traversability, vulnerability, victim
occupancy, robot occupancy, fire, and confidence values.
Conflicting reports are resolved by processing order,
without averaging or confidence-weighted fusion. Unreported values remain unchanged, while their confidence decreases by
$\sigma^{\mathrm{sup}}$, bounded below by zero, as in
\eqref{eqn:pconfijupdate}. The exploration layer is the cumulative union of the cells observed by all robots.

Additionally, the supervisor maintains a record containing each robot's latest reported position and current task. Its victim tracker
contains one record for each victim detected at least once,
storing the persistent index $v$, latest observed victim position, and supervisory status: \emph{roaming}, \emph{sought},
\emph{being rescued}, or \emph{rescued}. These statuses distinguish a detected victim awaiting assignment, an assigned victim whose robot is still searching, a victim under active guidance, and a completed rescue, respectively. Because binary victim occupancy does not encode identity,
persistent association of observations with victim index $v$ is assumed.

\noindent
\textbf{Task generation and allocation:} Exploration targets are extracted from frontiers, defined as explored,
perceived-traversable cells that are four-connected to an unexplored cell
represented as traversable. For each eight-connected frontier component of $n$ cells,
$m=\max\{1,\lfloor n/4\rfloor\}$ candidate targets are selected. For $m=1$, the frontier cell
nearest the component centroid is selected; otherwise, the frontier cells
nearest $m$ $k$-means cluster centers are selected, with duplicates
removed.

When rescue coordination is enabled, each newly detected victim
generates a rescue task at its observed position. Unallocated
rescue tasks are retained in a persistent backlog.

Each robot has fixed capability coefficients
$C_r^{\tau}\in[0,1]$. These coefficients are specified before the mission to represent
the robot's suitability for each task type. A zero coefficient excludes that task type,
whereas a positive coefficient determines its contribution to
the assignment utility.

Tasks are allocated through a modified sequential single-item auction \cite{b6}, assigning one eligible robot--task pair at a time according to the utility defined below.

For an auction triggered at step $k$, let
$\mathcal R_i^{\mathrm{available}}(k)$ and
$\mathcal T_i^{\mathrm{available}}(k)$ denote the participating
robots and unallocated tasks remaining at iteration $i$.
In addition to a type, each task $t$ has an associated target position
$\bqtask(k)$. For every eligible pair
$(r,t)\in\mathcal R_i^{\mathrm{available}}(k)
\times\mathcal T_i^{\mathrm{available}}(k)$, where task $t$ has type $\tau$, the assignment utility at iteration $i$ is

\begin{align}
\label{eqn:supervisory-utility}
\begin{split}
U_{r,t,i}^{\tau}(k)
&=
B^{\tau}C_{r}^{\tau}
\\ &-
\left\lVert
\bq(k)
-\bqtask(k)
\right\rVert_2
-
S_{t,i}(k).
\end{split}
\end{align}
Here, $B^{\tau}$ is the fixed-value reward associated with task type $\tau$. Rescue tasks use a larger task-type reward than exploration tasks so that victim assistance receives priority. The spatial-redundancy penalty $S_{t,i}(k)$ discourages assignments near existing exploration targets or unavailable
idle robots, as defined in
\eqref{eqn:supervisory-spatial-penalty}.

\begin{figure*}
\begin{equation}
\label{eqn:supervisory-spatial-penalty}
\small
S_{t,i}(k)
=
w^{\mathrm{task}}
\sum_{t^{\prime}\in \mathcal T_{i}^{\mathrm{explore}}(k)}
\exp\left(
-\frac{\norm{\bqtask(k) - \bqtaskprime(k)}_2}{\lambda^{\mathrm{s}}}
\right) +
w^{\mathrm{robot}}
\sum_{r^{\prime}\in \mathcal R_{i}^{\mathrm{idle}}(k)}
\exp\left(
-\frac{\norm{\bqtask(k) - \bqprime}_2}{\lambda^{\mathrm{s}}}
\right).
\end{equation}
\end{figure*}

In \eqref{eqn:supervisory-spatial-penalty}, $\mathcal T_i^{\mathrm{explore}}(k)$ contains exploration
tasks assigned to robots not participating in the current auction
iteration, while $\mathcal R_i^{\mathrm{idle}}(k)$ contains idle
robots not participating in that iteration. The first sum penalizes proximity between the
candidate target $\bqtask(k)$ and existing exploration targets
$\bqtaskprime(k)$. The second penalizes proximity to the latest
reported positions $\bqprime(k)$ of unavailable idle robots.
Both penalties decrease with distance, with non-negative weights
$w^{\mathrm{task}}$ and $w^{\mathrm{robot}}$ and decay length
$\lambda^{\mathrm{s}}>0$.

Robots executing rescue tasks are excluded because this penalty
represents overlapping exploration effort, rather than general
collision avoidance. Proximity to an ongoing rescue therefore
does not by itself discourage a nearby assignment.

Before finalizing the highest-utility robot--task pair, the supervisory layer uses four-connected \gls{bfs} on its shared belief to find a
minimum-length path from the robot's latest reported position to the task
target. Intermediate path cells must be perceived as traversable,
victim-free, and non-burning; perceived robot occupancy is ignored. The
target cell is exempt from these conditions, allowing a rescue path to
terminate at a victim, although an obstructed or burning target may
subsequently be impossible to enter.

If no path exists, the task is removed from the current auction while
the robot remains available. Unreachable rescue tasks remain in the
persistent backlog, whereas unreachable exploration tasks are discarded. Otherwise, the allocation is finalized and the ordered path
$\Pi_r(k)=
(\boldsymbol q_{r,0}^{\mathrm{path}},\ldots,
\boldsymbol q_{r,\ell}^{\mathrm{path}})$
is transmitted with the task. Here, $\ell$ is the number of
cardinal movements along the path, whose consecutive cells
are four-connected, starting at the robot's reported position
and ending at the task target. The
robot and task are then removed from the current auction, and an allocated
rescue task is removed from the backlog. This procedure continues until
no eligible pair remains.

\noindent
\textbf{Event-triggered intervention:} The supervisor processes the following triggered events:
\begin{itemize}
    \item \emph{Spatial conflict} when a robot perceives another robot within
    its sensing radius $\rrobot$. To prevent repeated reports of the same
    persistent encounter, further spatial-conflict reports from that robot
    are suppressed for $n^{\mathrm{conflict}}$ steps.

    \item \emph{Low progress} when an idle robot repeatedly contributes no new cells to the shared exploration map. The supervisor maintains a separate impatience counter for each robot, incrementing it after such a scan and resetting it when new cells are contributed or the robot is no longer idle. A low-progress event is generated when the counter exceeds $n^{\mathrm{impatience}}$, after which it is reset.

    \item \emph{Blocked path} where during an exploration task, an upcoming
    cell on $\Pi_r(k)$ is no longer perceived as traversable,
    victim-free, robot-free, or non-burning. The robot then becomes idle.

    \item \emph{Victim discovery} when a scan detects a victim whose persistent index is not yet recorded in the victim tracker. The supervisor creates a record for that victim.

    \item \emph{Task life-cycle event} when an exploration task is completed,
    active guidance begins or is abandoned, or a victim is rescued
    or dies.
\end{itemize}

Victim-discovery and task life-cycle events are processed
immediately. A newly generated rescue task opens the auction
to assignable robots not already performing rescue. Spatial-conflict, low-progress, and
blocked-path events request reallocation and regeneration of exploration
targets, but intervention occurs only after $n^{\mathrm{event}}$ unresolved events of
these types have accumulated. 

\begin{table}[t]
    \centering
    \caption{Task-aware control and rescue-guidance notation.}
    \label{tab:taskawarevariables}
    \begin{tabular*}{\columnwidth}{
        @{\extracolsep{\fill}}lp{0.54\columnwidth}@{}
    }
        \toprule
        \textbf{Variable} & \textbf{Description} \\
        \midrule
        $L_r^{\mathrm{supervisor}}(k^{\prime}\mid k)$
        & Task-aware stage objective \\

        $\phiprox(k^{\prime}\mid k)$
        & Robot-proximity score \\

        $\phipath(k^{\prime}\mid k)$
        & Path-progress score \\

        $\mathcal Q_r^{\mathrm r}(k^{\prime}\mid k)$
        & Perceived positions of other robots \\

        $j_r^\star(k^{\prime}\mid k)$
        & Nearest path-cell index \\

        $d_r^\star(k^{\prime}\mid k)$
        & Distance to nearest path cell \\

        $s_r^{\mathrm{guide}}(k)$
        & Robot guidance score \\

        $d_r^{\mathrm{rv}}(k)$
        & Robot-victim distance \\

        $\Delta d_r^{\mathrm{rv}}(k)$
        & Modified separation-decrease signal \\
        \bottomrule
    \end{tabular*}
\end{table}

\noindent
\textbf{Task-aware local control:} The task-aware controller retains the optimization structure of \eqref{eqn:mpc-objective}, but replaces its stage objective with \eqref{eqn:task-aware-stage-objective}. Idle robots prioritize discovery, whereas assigned exploration
and rescue search prioritize progress along $\Pi_r(k)$. In \eqref{eqn:task-aware-stage-objective}, all weights are non-negative
tuning parameters. The discovery score $\phidisc$, confidence score $\phiconf$, and vulnerability score $\phivuln$ are defined in
Table~\ref{tab:mpc-objective-components}, whereas the proximity score $\phiprox$ and path-progress score $\phipath$ are defined in Table~\ref{tab:mpc-objective-components-prox-path}.

While the robot is idle, the terminal exploration contribution in
\eqref{eqn:mpc-terminal-objective} remains active. It is set to zero
during an exploration task or the searching stage of a rescue task
because the supervisor-provided path supplies the long-range directional
objective. No terminal \gls{mpc} contribution is evaluated during active
guidance.

For a non-wait candidate movement, the task-aware \gls{mpc} controller treats
a destination currently perceived to be occupied by another robot as
infeasible. This supplements the common perceived-feasibility conditions
from Section~\ref{subsubsec:modelpredictivecontrollers}. Conflicting commands arising from incomplete or outdated
perceptions remain subject to the movement-execution rules
in Section~\ref{subsec:common-experimental-setup}.

\begin{figure*}
\begin{equation}
\small
\label{eqn:task-aware-stage-objective}
L_r^{\mathrm{supervisor}}(k^{\prime}\mid k)
=
\begin{cases}
\begin{aligned}
&\omegadisc
 \phidisc(k^{\prime}\mid k)
+
\omegaconf
 \phiconf(k^{\prime}\mid k)
\\
&
-
\omegavuln
 \phivuln(k^{\prime}\mid k)
-
\omegaprox
 \phiprox(k^{\prime}\mid k),
\end{aligned}
&
\text{if robot is idle}
\\ \\[1ex]
\begin{aligned}
&\omegapath
 \phipath(k^{\prime}\mid k)
+
\omegaconftilde
 \phiconf(k^{\prime}\mid k)
\\
&
-
\omegavulntilde
 \phivuln(k^{\prime}\mid k)
-
\omegaproxtilde
 \phiprox(k^{\prime}\mid k),
\end{aligned}
&
\begin{aligned}
&\text{if }\tasktype(k)=\mathrm{explore}, \text{ or}\\
&\text{if }\tasktype(k)=\mathrm{rescue} \text{ and} \\
&\taskstatus(k)=\mathrm{searching}.
\end{aligned}
\end{cases}
\end{equation}
\end{figure*}

\begin{table*}[t]
\centering
\caption{Robot-proximity and path-progress components of the \gls{mpc} stage
objective. The set $\mathcal Q_r^{\mathrm{r}}(k^{\prime}\mid k)$ contains the
perceived positions of all robots other than $r$, and $\psi(d)$ is a proximity penalty that is non-increasing
for positive Euclidean separation $d>0$. Its numerical definition is given in Table~\ref{tab:model-variables} in Appendix~\ref{app:configurations}.
For the path $\Pi_r(k)$ defined above,
$j_r^\star(k^{\prime}\mid k)$ is the zero-based index of the path cell nearest
to $\bq(k^{\prime}\mid k)$ among the first
$\min\{n^{\mathrm{path}},\ell+1\}$ remaining path cells, and
$d_r^\star(k^{\prime}\mid k)$ is the corresponding Euclidean distance.
The constants $\beta_0^{\mathrm{path}}$, $\beta_1^{\mathrm{path}}$, and $\beta_2^{\mathrm{path}}$ respectively determine
the baseline score, reward for forward path progress, and penalty for
deviation from the path.}
\label{tab:mpc-objective-components-prox-path}
\begingroup
\setlength{\tabcolsep}{3pt}
\renewcommand{\arraystretch}{1.45}

\begin{tabularx}{\textwidth}{
    @{}
    >{\raggedright\arraybackslash}p{0.20\textwidth}
    >{\raggedright\arraybackslash}X
    @{}
}
\toprule
\textbf{Component} & \textbf{Definition} \\
\midrule

Robot proximity
&
\(
\begin{aligned}
\phiprox(k^{\prime}\mid k)
&=
\min\Biggl\{
1,\,
\sum_{\boldsymbol c\in
\mathcal Q_r^{\mathrm r}(k^{\prime}\mid k)}
\psi \left(
\left\|
\bq(k^{\prime}\mid k)-\boldsymbol c
\right\|_2
\right)
\Biggr\}
\end{aligned}
\)
\\

Path progress
&
\(
\begin{aligned}
\phipath(k^{\prime}\mid k)
&=
\operatorname{clip} \Bigl(
\beta_0^{\mathrm{path}}
+\beta_1^{\mathrm{path}}j_r^\star(k^{\prime}\mid k)
-\beta_2^{\mathrm{path}}d_r^\star(k^{\prime}\mid k),
\,0,1
\Bigr)
\end{aligned}
\)
\\

\bottomrule
\end{tabularx}

\endgroup
\end{table*}

\noindent
\textbf{Trust-aware active rescue guidance:} A rescue task changes from searching to active guidance when
the robot satisfies a proximity criterion relative to its
stored rescue target. The distance-based criterion requires their Euclidean separation to be at most $d^{\mathrm{init}}$. Upon activation, the rescue-guidance policy replaces the task-aware \gls{mpc} controller.

During active guidance, robot $r$ maintains a guidance score
$s_r^{\mathrm{guide}}(k)\in[0,1]$ for its assigned victim $v$.
Let $d_r^{\mathrm{rv}}(k)=\|\bq(k)-\bqv(k)\|_2$
denote their ground-truth separation. The update signal
$\Delta d_r^{\mathrm{rv}}(k)$ equals
$d_r^{\mathrm{rv}}(k-1)-d_r^{\mathrm{rv}}(k)$, except that unchanged separation, or increasing separation when the
preceding distance was within the close-range threshold, yields
$\Delta d_r^{\mathrm{rv}}(k)=\varepsilon^{\mathrm{guide}}>0$.
After the first active-guidance step,
\begin{align}
\begin{split}
\label{eqn:robot-trust-estimate}
s_r^{\mathrm{guide}}(k)
=
\operatorname{clip}\big(
&s_r^{\mathrm{guide}}(k-1)
\\
&+
\sigma^{\mathrm{guide}}\Delta d_r^{\mathrm{rv}}(k),
0,1
\big),
\end{split}
\end{align}
where $\sigma^{\mathrm{guide}}>0$ is the update gain.
The first active-guidance step initializes the stored distance
without changing the score. Guidance uses the victim's
ground-truth position, including outside the robot's sensing
region. The victim's decision to follow remains governed
independently by its trust $\theta_v(k)$ in
\eqref{eqn:victim-trust-update}.

The guidance score selects between approach and lead policies.
Both use an exit-directed action that selects an in-grid, four-connected, perceived-traversable cell with the smallest distance to the
exit. These distances are recomputed at each step by
four-connected \gls{bfs} using only the robot's perceived
traversability layer $\ptrav$.

When $s_r^{\mathrm{guide}}(k)<\theta^{\mathrm{lead}}$,
the approach policy first attempts to bring the robot within the victim's
sensing radius. Outside that radius, it selects an in-grid,
perceived-traversable neighbor minimizing the Manhattan
distance to the victim's ground-truth position; inside it,
the robot takes an exit-directed action.

When $s_r^{\mathrm{guide}}(k)\geq\theta^{\mathrm{lead}}$,
the lead policy takes an exit-directed action while the victim
remains within the leash distance $d^{\mathrm{leash}}$;
otherwise, the robot waits. The policy and distance conditions
are reassessed at every step.

Active guidance is abandoned if the robot exceeds a leash-wait or no-progress limit specified in Table~\ref{tab:model-variables}. The robot returns to idle and reports the abandonment to the supervisor, and if the victim remains tracked, the rescue task
returns to the backlog. Rescue completion is reported when the assigned victim enters the predefined exit region.

\section{Experimental evaluation}
\label{sec:experiments}

This section evaluates the proposed architecture through complementary
studies. We first compare the \gls{mpc}-based controllers and examine
tube adaptation and sensitivity to measurement noise. We then assess
reward design and recurrent memory for learned exploration, before
comparing model-based, learned, and shielded controllers in a common
setting. Finally, we evaluate supervisory coordination with homogeneous
and heterogeneous teams, followed by rescue-team sizing and adaptive
victim guidance.

\subsection{Common environment and evaluation procedure}
\label{subsec:common-experimental-setup}

The experiments use a discrete-time simulator implementing
the models and decision rules introduced in
Section~\ref{sec:problem-formulation}. The simulator maintains the
ground-truth environment and updates robot motion, victim
behavior, and environmental dynamics.

The evaluation comprises five scenario families and eight base environment configurations: $\mathcal E^{\mathrm{MPC}}$ for local \gls{mpc} comparisons and measurement-noise tests,
$\mathcal E^{\mathrm{RLD}}$ for \gls{rl} development at three map scales,
$\mathcal E^{\mathrm{RLE}}$ for memory and final-controller
comparisons, $\mathcal E^{\mathrm{SV}}$ for homogeneous and
heterogeneous supervisory studies, and $\mathcal E^{\mathrm{DR}}$
for dynamic rescue. Study-specific settings are introduced below, and additional parameters and results appear in Appendices~\ref{app:configurations}
and~\ref{app:supplementary-results}, respectively.

Map dimensions vary across these complementary studies:
the larger supervisory environments accommodate distributed
search by multiple robots, while \gls{rl}-development study explicitly varies map size to assess exploration as the search space grows.

Each simulated map consists of rooms connected by corridors, generated procedurally using
the dungeon-style map-generation approach of \cite{b18}.
For the vulnerability layer defined in
Section~\ref{subsec:environment}, the experiments use
$\nusafe=0$, $\numoderate=0.6$, and $\nuhigh=1$, adopting the numerical levels of the normalized combustibility classification in \cite{b36} as terrain-risk values.

All cells initially have zero vulnerability. Each corridor and room is independently selected for vulnerability assignment with a configured corridor- or room-specific probability. For each selected region, a second random draw determines whether all its cells receive the moderately vulnerable or high-risk value, using the corresponding conditional high-risk probability. These assignments are
applied first to corridor cells and then to room cells, so a selected room's vulnerability value takes precedence at cells belonging to both regions.

Victims are placed by uniformly sampling a room and then an
unoccupied cell within it. Initial fire locations are sampled
similarly, excluding victim-occupied cells and previously
selected fire locations. These cells are initialized as burning,
with eligible adjacent cells marked as flammable, following
Section~\ref{subsec:environment}. Comparative evaluations use matched map seeds, giving methods
the same initial geometry, terrain-risk distribution, victim placement, and fires.

Victim treatment distinguishes detection from evacuation.
In $\mathcal E^{\mathrm{MPC}}$, $\mathcal E^{\mathrm{RLD}}$,
and $\mathcal E^{\mathrm{RLE}}$, victims are stationary
detection targets and motion obstacles, and neither detection nor
fire removes them. In $\mathcal E^{\mathrm{SV}}$ and
$\mathcal E^{\mathrm{DR}}$, victims occupying burning or burnt
cells are removed as fatalities after the victim-movement
stage and before fire propagation. In homogeneous supervisory missions, victims remain stationary.
Detection completion requires at least one surviving victim and detection of every survivor. The heterogeneous supervisory and dynamic-rescue studies
additionally evaluate evacuation, with victims considered
rescued upon reaching an exit. 

All robots scan the environment at $k=0$ before selecting their
first command. Each subsequent step processes robot commands and
movement conflicts, victim movement where enabled, stochastic
displacements, and a fire update, followed by automatic scanning.
Robot scans update the local perceptions used for control, while dynamic victims scan their surroundings to update the perceptions used in their trust and movement rules.

In $\mathcal E^{\mathrm{DR}}$, an additional fire update follows
scanning, allowing newly observed fire information to become
outdated before the next decision. This represents an
observation--action mismatch relevant to evacuation planning
with time-dependent environmental knowledge \cite{b42}.
Both fire spread and burning duration therefore advance twice per step.

Controller constraints and safety checks use perceived
conditions, whereas movement execution uses the ground-truth
environment. However, incomplete perception, stochastic displacements, or
conflicting robot commands can still produce an invalid
destination. Furthermore, unshielded \gls{rl} lacks these feasibility checks.

Irrespective of the controller, commanded movements and stochastic displacements directed outside the grid or into an obstruction (terrain collisions), or into a victim-occupied cell (victim collisions), are blocked and recorded. Blocking a command preserves the current position, whereas blocking a displacement
preserves the post-command position. Conflicting robot commands are also blocked and recorded as inter-robot collisions.

Fire collisions are recorded when a robot occupies a burning
cell after command processing, before stochastic displacements
and fire propagation. Burning cells do not themselves block
movement execution. This fire-collision counter is absent from
the implementation used for $\mathcal E^{\mathrm{RLD}}$ and
$\mathcal E^{\mathrm{RLE}}$, including its model-based baselines.

Each recorded collision contributes one harm event. To represent harm from vulnerable terrain beyond recorded
collisions, each robot incurs one additional harm event with probability equal to the ground-truth vulnerability of its post-disturbance cell at each step. This is a modeling convention: physical contact, impact dynamics, and degradation of robot capabilities are not simulated.

Coverage is the fraction of traversable
cells observed. In $\mathcal E^{\mathrm{MPC}}$,
$\mathcal E^{\mathrm{RLD}}$, and $\mathcal E^{\mathrm{RLE}}$,
recall is the fraction of initially placed victims detected,
and success requires detecting all of them before the deadline. Homogeneous supervisory results use the detection-completion
condition specified above, and rescue studies report victims
rescued, deceased, and remaining. Coverage, recall, and
completion rates are expressed as percentages, and all
\glspl{ci} are at the $95\%$ level.

\subsection{\gls{mpc}-based controllers}
\label{subsec:local-control-experiments}

The primary comparison evaluates nominal \gls{mpc}, fixed-tube
\gls{tmpc}, and \gls{sddtmpc} on 500 matched $35\times35$ maps
containing 12 victims and two initial fires. A single robot is given
300 steps. We report both \emph{completion}, which ignores collisions, and \emph{collision-free completion}, which additionally
requires no recorded terrain, victim, or fire collision. Completion-time
comparisons exclude timeouts and use only seeds completed by both
controllers in each pair. The perceived safety margin is the Manhattan distance from the
controller's selected target cell to the nearest perceived boundary,
wall, victim, or fire hazard, and is set to zero when no feasible
action is available. Its run average includes every simulation
step and is evaluated over the same 500 matched seeds.

A separate study evaluates \gls{sddtmpc} on 250 matched maps, varying the single robot's vulnerability-measurement noise bound over $\deltavuln\in\{0,0.1,0.2,0.4\}$ while retaining the same
map size, victim count, and time limit. Here, collision-free completion excludes terrain and victim collisions only. Fire collisions are reported separately and do not affect this criterion. 

\subsubsection{Safety, completion, and mission duration}
\label{subsec:mpcsafetycompletionduration}

\begin{figure}[t]
    \centering
    \includegraphics[width=0.9\linewidth]{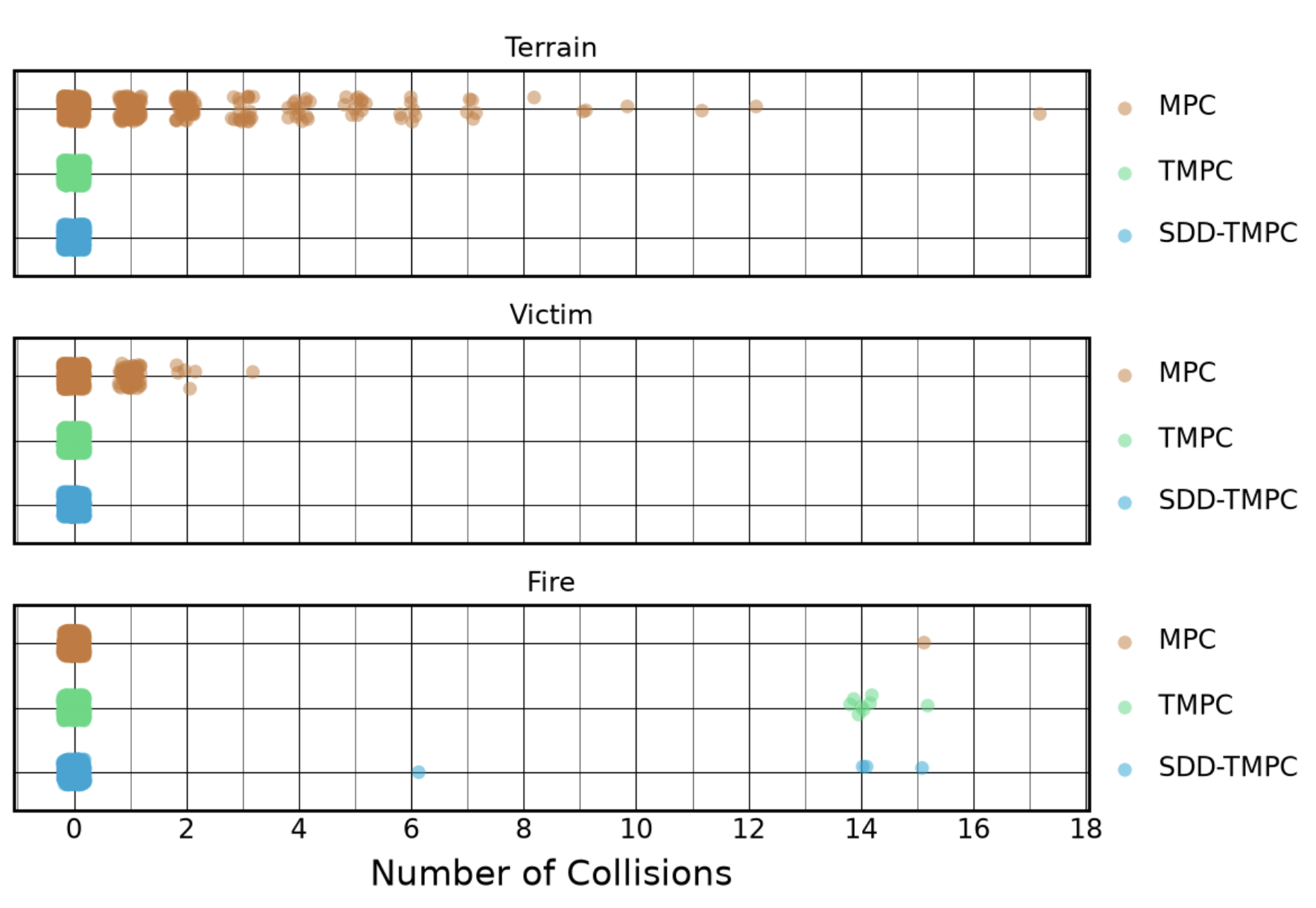}
    \caption{Recorded collisions for the \gls{mpc}-based controllers per run in the 500-seed local-control
    comparison.}
    \label{fig:lc-collisions}
\end{figure}

The 500-seed comparison first assesses whether explicit protection against displacements reduces collisions. Neither tube-based controller records terrain or
victim collisions (Figure~\ref{fig:lc-collisions}), whereas nominal \gls{mpc} records at least one terrain,
victim, or fire collision in 240 runs ($48.0\%$, Wilson \gls{ci} $[43.7\%,52.4\%]$), with up to
17 terrain collisions in one run. 

Fire remains a limitation: fire-collision counts differ across the
three controllers (Friedman $\chi^2=7.40$, $p=0.025$). Wilcoxon signed-rank tests, Bonferroni-adjusted across the
three controller pairs, identify more fire collisions for
fixed-tube \gls{tmpc} than nominal \gls{mpc} ($p=0.024$),
but no significant difference involving \gls{sddtmpc}.  Thus, the observed protection against terrain and victim
collisions provided by fixed-tube \gls{tmpc} and
\gls{sddtmpc} does not extend to eliminating fire collisions. One possible but unverified mechanism arises when a robot's current cell becomes burning: a tube-based controller may reject an otherwise safe escape move because a possible displacement would return the robot to that cell. If no feasible escape remains, the resulting wait prolongs fire exposure.

Table~\ref{tab:lc-success} shows the corresponding completion trade-off.
Nominal \gls{mpc} finishes most often, but its completion rate falls
from $97.2\%$ to $50.2\%$ when collisions count as failure.
\Gls{sddtmpc} retains $81.2\%$ collision-free completion, compared
with $34.4\%$ for  fixed-tube \gls{tmpc}. All paired controller comparisons
are significant under both criteria (exact McNemar tests, $p<0.001$).
Adaptive tightening therefore achieves substantially more
collision-free completions than either nominal planning or
the fixed tube under the tested conditions.

\begin{table}[t]
    \centering
    \caption{Completion rates (\%) over 500 matched runs.
    Collision-free completion excludes all three recorded collision types.}
    \label{tab:lc-success}
    \begin{tabular*}{\columnwidth}{@{\extracolsep{\fill}}lcc@{}}
        \toprule
        \textbf{Controller} & \textbf{Completion} & \textbf{Collision-free} \\
        \midrule
        \gls{mpc} & 97.2 & 50.2 \\
        \gls{tmpc} & 35.0 & 34.4 \\
        \gls{sddtmpc} & 81.6 & 81.2 \\
        \bottomrule
    \end{tabular*}
\end{table}

\begin{figure}[t]
    \centering
    \includegraphics[width=0.9\linewidth]{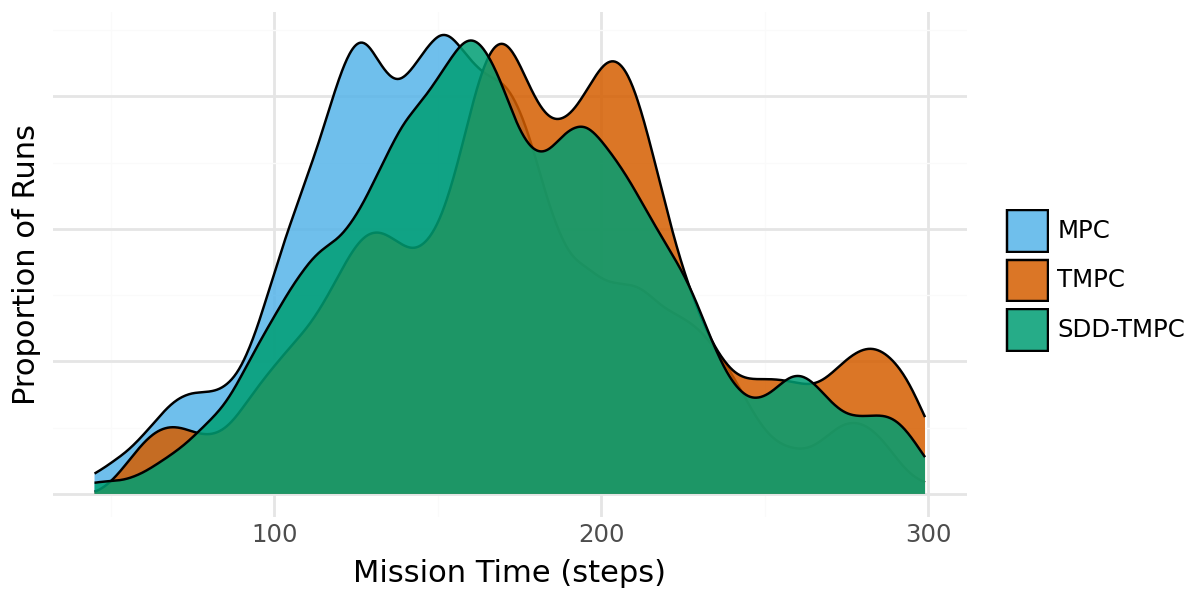}
    \caption{Kernel-density estimates of successful mission durations.
    Timeouts are excluded and each controller's density is normalized
    separately. Small tail peaks are sensitive to smoothing.}
    \label{fig:lc-durations}
\end{figure}

For seeds completed by both controllers, \gls{sddtmpc}
finishes an average of 13 steps earlier than fixed-tube
\gls{tmpc} and 14 steps later than nominal \gls{mpc}
(Wilcoxon signed-rank tests, Bonferroni-adjusted across
the three controller pairs, $p<0.001$). Thus, adaptation improves speed as well as completion
relative to a fixed tube. Figure~\ref{fig:lc-durations} shows the successful-run duration
distributions. 

\subsubsection{Sensitivity to vulnerability-measurement noise}
\label{subsec:mpcnoisesensitivity}

We next examine whether the safety and completion benefits of
\gls{sddtmpc} persist as vulnerability-measurement noise increases. No terrain or victim collisions are observed at the tested noise
bounds $\deltavuln\in\{0,0.1,0.2\}$, whereas at
$\deltavuln=0.4$ they occur in nine of 250 runs
($3.6\%$, Wilson \gls{ci} $[1.9\%,6.7\%]$).
The paired comparison between noise bounds $0.2$ and $0.4$
confirms this increase in collision incidence
(exact McNemar $p=0.0039$).
A possible mechanism is vulnerability underestimation: when perceived
vulnerability falls below confidence, the controller deactivates its
tube, although the ground-truth vulnerability may still cause a displacement.
The observed collision protection therefore depends on sensing
accuracy and is lost at the largest tested noise bound.

Table~\ref{tab:lc-noise-success} distinguishes completion from
completion without terrain or victim collisions. The latter varies
across the three non-zero noise levels
(Cochran's $Q=11.61$, $p=0.0030$), decreasing from $84.8\%$ for 
$\deltavuln=0.1$ to $75.2\%$ for 
$\deltavuln=0.4$.
For comparison, fixed-tube \gls{tmpc}, evaluated without measurement
noise on the same 250 seeds, achieves only $39.2\%$ for the 
collision-free completion rate, which is lower than that of \gls{sddtmpc} at every
tested noise level (exact paired McNemar tests, all $p<0.001$).
Thus, increasing noise compromises collision protection without
eliminating the completion advantage over the conservative fixed-tube \gls{tmpc}.
The small completion-rate increase for $\deltavuln = 0.1$ does not
establish a benefit from noisier sensing.

\begin{table}[t]
    \centering
    \small
    \caption{Uncertainty-study outcomes over 250 matched runs per configuration. All outcome columns are percentages.
    Collision-free completion excludes terrain and victim collisions,
    but not fire collisions. The final column gives the percentage
    of runs containing at least one terrain or victim collision.
    The fixed-tube reference is evaluated without measurement noise.}
    \label{tab:lc-noise-success}
    \begin{tabular*}{\columnwidth}{
    @{\extracolsep{\fill}}
    cccc
    @{}
    }
    \toprule
    $\deltavuln$
    & \textbf{Completion}
    & \makecell{\textbf{Collision-free}}
    & \makecell{\textbf{Runs with}\\\textbf{a collision}}
    \\
    \midrule
    
    \multicolumn{4}{@{}l}{\textbf{\gls{sddtmpc}}} \\
    0.0 & 82.4 & 82.4 & 0.0 \\
    0.1 & 84.8 & 84.8 & 0.0 \\
    0.2 & 78.8 & 78.8 & 0.0 \\
    0.4 & 76.8 & 75.2 & 3.6 \\
    
    \midrule
    \multicolumn{4}{@{}l}{\textbf{\gls{tmpc}}} \\
    0.0 & 39.2 & 39.2 & 0.0 \\
    
    \bottomrule
    \end{tabular*}
\end{table}

\subsubsection{Tube adaptation and trajectory behavior}
\label{subsec:mpctubeadaptation}

\begin{figure}[t]
    \centering
    \includegraphics[width=\linewidth]{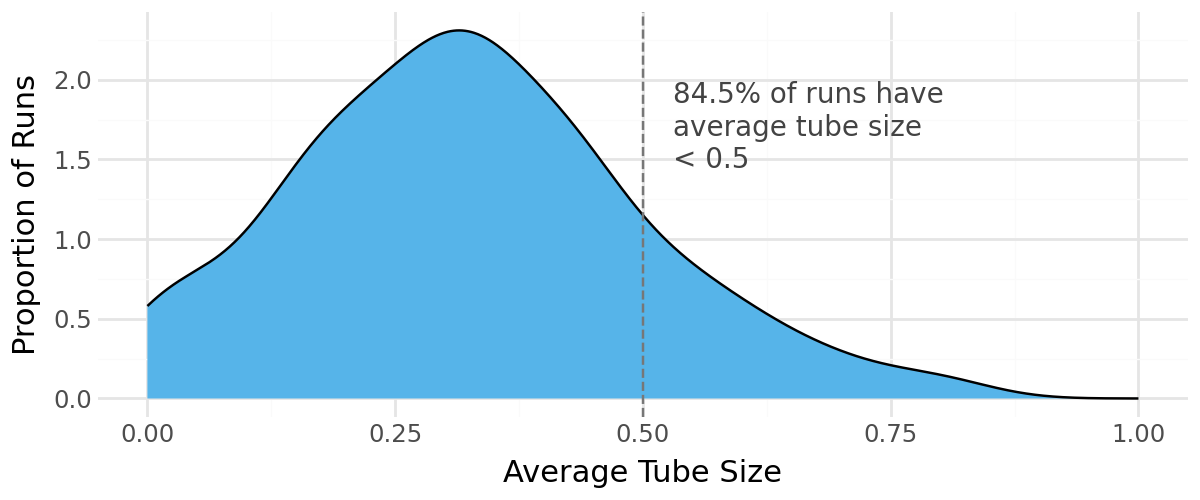}
    \caption{Kernel-density estimate of the run-average tube-activation
    fraction over 206 successful noise-free runs. Zero indicates no
    activation at selected command destinations, whereas one indicates activation
    at every step. The dashed line marks an activation fraction of $0.5$,
    below which $84.5\%$ of runs fall.}
    \label{fig:lc-tube-density}
\end{figure}

For \gls{sddtmpc}, the recorded tube-activation fraction
measures how often the selected command destination
activates the one-cell tube constraints. Across the 206 successful noise-free runs in the 250-seed uncertainty study, its mean is $0.323$ (bootstrap \gls{ci} $[0.299,0.347]$), and $84.5\%$ of runs have a fraction below $0.5$ (Figure~\ref{fig:lc-tube-density}).

Among these successful runs, an exploratory analysis detects no monotonic association between activation fraction and completion time (Spearman correlation $-0.012$; two-sided permutation $p=0.866$). The 44 timed-out runs exhibit a higher mean activation fraction of $0.452$.

\begin{figure}[t]
    \centering
    \begin{subfigure}[b]{0.32\linewidth}
        \includegraphics[width=\textwidth]{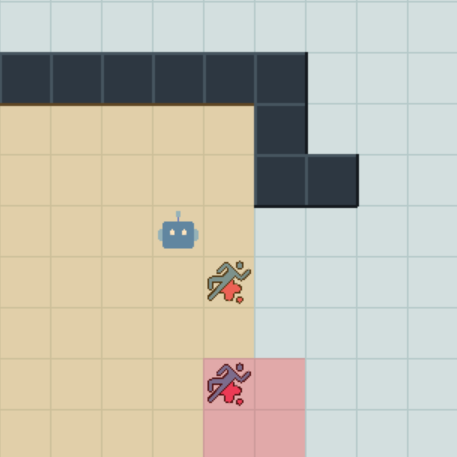}
        \caption{Step 69}
    \end{subfigure}
    \begin{subfigure}[b]{0.32\linewidth}
        \includegraphics[width=\textwidth]{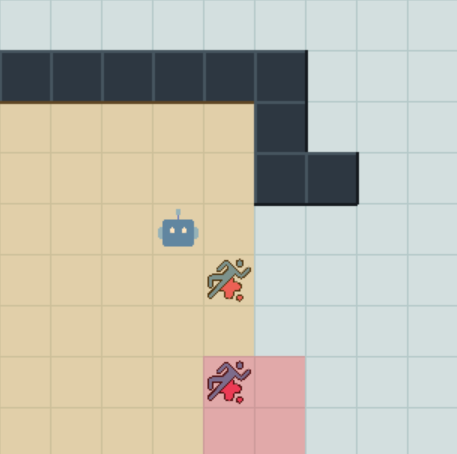}
        \caption{Step 70}
    \end{subfigure}
    \begin{subfigure}[b]{0.32\linewidth}
        \includegraphics[width=\textwidth]{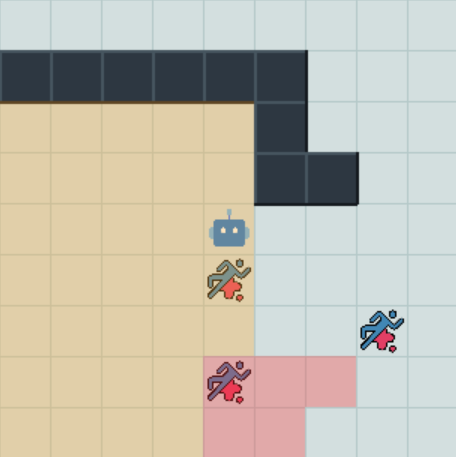}
        \caption{Step 71}
    \end{subfigure}
    \caption{Robot perception at steps 69--71 during an
    \gls{sddtmpc} run. Dark cells denote walls, pale blue-grey
    cells safe terrain, yellow cells moderately vulnerable terrain,
    and red cells high-risk terrain. Icons identify robots and
    victims.}
    \label{fig:lc-confidence-passage}
\end{figure}

Figure~\ref{fig:lc-confidence-passage} illustrates how repeated
observations can make a narrow passage feasible. \gls{sddtmpc} initially
waits because confidence is insufficient to certify a one-cell passage
beside a victim. After confidence increases, the estimated displacement probability becomes
zero and the robot traverses the gap. Nominal constraints permit
entry without this confidence, whereas the fixed tube excludes the passage.

A binary logistic regression over the 500 matched seeds
relates the occurrence of any nominal-\gls{mpc} terrain,
victim, or fire collision to the timeout indicators of both
tube-based controllers. Adjusting for fixed-tube timeouts,
\gls{sddtmpc} timeouts are associated with higher collision
odds under nominal \gls{mpc} (odds ratio $3.54$,
\gls{ci} $[2.09,5.97]$, $p<0.001$). The fixed-tube timeout indicator shows no corresponding
association ($p=0.716$). The association between
\gls{sddtmpc} timeouts and nominal-\gls{mpc} collisions
does not establish that those scenarios lack a collision-free route.

\begin{figure}[t]
    \centering
    \includegraphics[width=\linewidth]{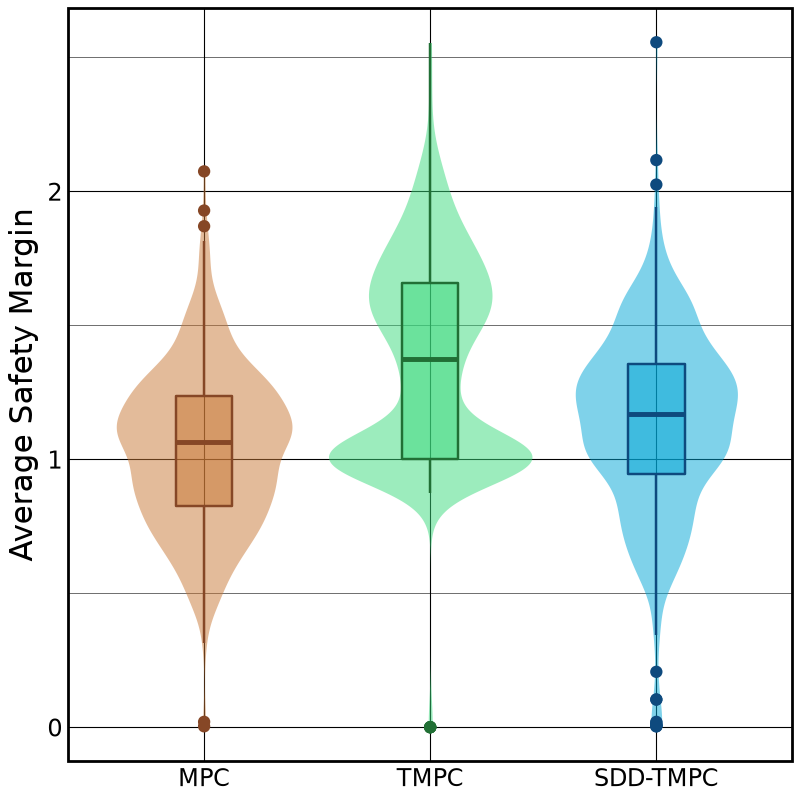}
    \caption{Distribution of run-average perceived safety margins over
    500 matched runs per controller. Margins measure Manhattan clearance
    at the target cell, with zero assigned when no feasible action is available. All steps, including those in timed-out runs,
    contribute to the averages.}
    \label{fig:lc-safety-margin}
\end{figure}

Finally, fixed-tube \gls{tmpc} has the largest run-average perceived safety
margin, nominal \gls{mpc} the smallest, and \gls{sddtmpc} lies between
them (Figure~\ref{fig:lc-safety-margin}). The controllers differ overall
(Friedman $\chi^2=380.1$, $p<0.001$), with differences between every
pair (Bonferroni-adjusted Wilcoxon signed-rank tests, $p<0.001$).

Within \gls{pasar}, these results support adaptive tightening as a compromise between perceived hazard clearance and search progress. The larger margins of fixed-tube \gls{tmpc} do not compensate for its substantially lower completion rate.

\subsection{\gls{rl}-based controllers}
\label{subsec:rl-experiments}

We examine how reward design and observation memory affect
learned exploration using \gls{ppo}. Reward and sensitivity
studies first assess a policy receiving four stacked observations,
followed by a held-out comparison with the recurrent policy of
Section~\ref{subsubsec:rl-controller}. Network architectures,
training settings, and experimental variants are listed in
Table~\ref{tab:model-variables},
Appendix~\ref{app:configurations}.

During reward development, inactive behavior was observed
with a preliminary design that directly penalized vulnerability and fire exposure. A second design removed these penalties and rewarded the first physical visit to each cell, while retaining a terminal timeout penalty. This penalty provides no distinction between
partial exploration and little progress when neither leads to completion.

The final reward in \eqref{eqn:rl-reward} therefore emphasizes exploration progress through coverage-potential shaping and terminal coverage, replacing the first-visit bonus and omitting the timeout penalty. Detection and completion rewards and step and collision penalties are retained.

Using the final reward in \eqref{eqn:rl-reward} as the reference, the sensitivity study varies discounting, initial-position randomization, episode length, potential shaping, and training duration. The shaping variant replaces the coverage potential while retaining the reward's overall structure.

Separate policies are trained on $20\times20$, $25\times25$, and $30\times30$ maps. Larger-map configurations also increase the victim count, episode limit, and training budget (Table~\ref{tab:model-variables}), assessing increasingly
demanding configurations. Coverage ranges from $56\%$ to $62\%$ across the tested $20\times20$ configurations and reaches $41.5\%$ in the $30\times30$ configuration (Table~\ref{tab:rl-sensitivity}).

These single-training-seed results, measured on final training rollouts, suggest that the tested parameter and budget changes do not resolve the exploration limitation. One possible constraint is the four-frame observation window, which provides only a short temporal history beyond the accumulated information stored in the perception maps. We therefore test
whether recurrent memory improves exploration, without assuming
that limited memory alone explains the development results.

\begin{figure}[t]
    \centering
    \includegraphics[width=\linewidth]{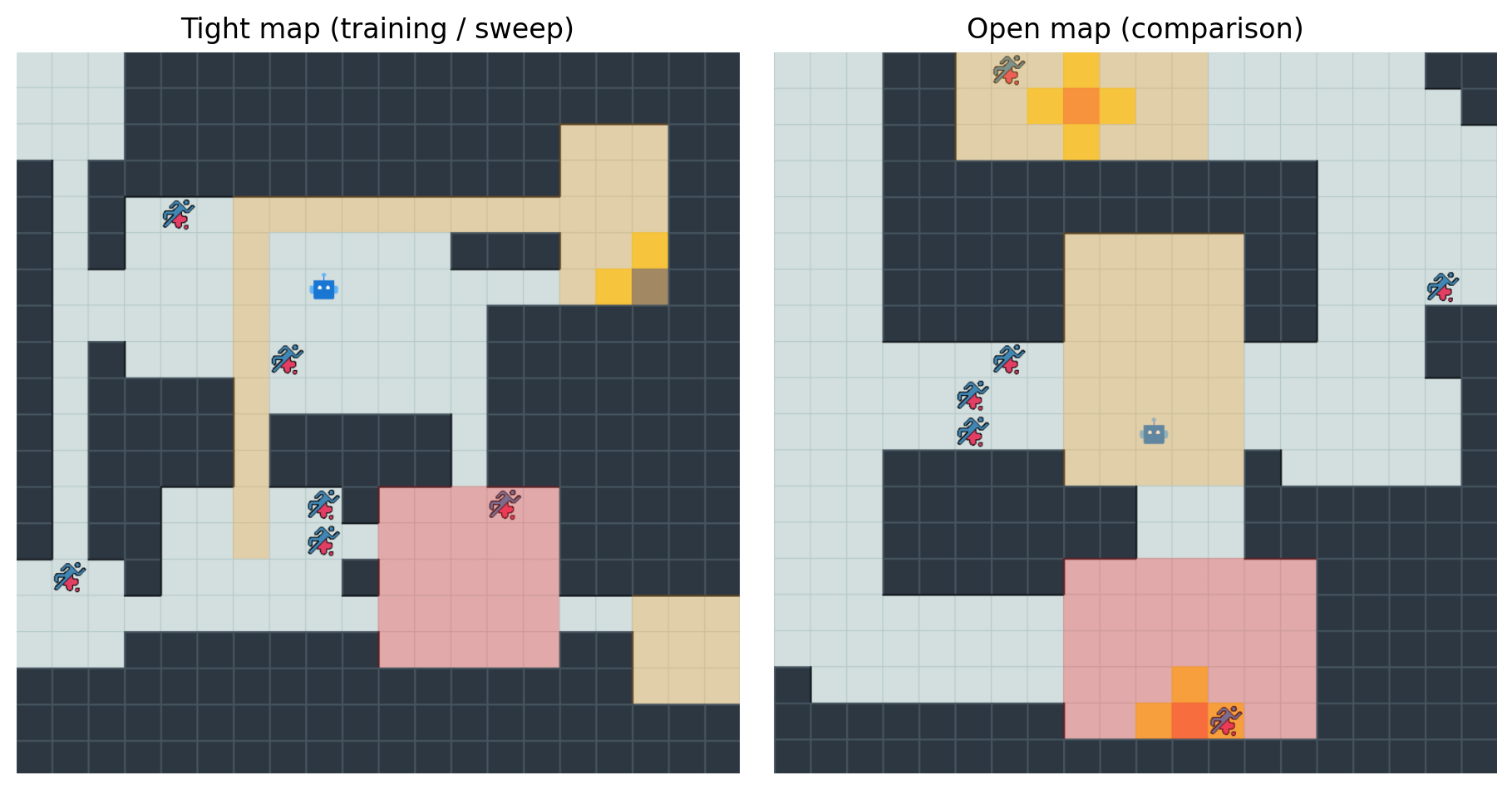}
    \caption{Representative tight and open environment configurations.
    The open-map configuration is shared by all five controllers in
    the final comparison and is also used for the memory comparison.}
    \label{fig:rl-maps}
\end{figure}

The memory comparison uses $\mathcal E^{\mathrm{RLE}}$:
$20\times20$ open maps with six stationary victims, two
initially burning cells, three-cell-wide corridors, and
room-based vulnerability (Figure~\ref{fig:rl-maps}).
This tests whether recurrent memory improves exploration
even at the smallest examined map size.
Training uses random initial positions and a 1000-step
episode limit; held-out evaluation starts from the map
center and uses a 500-step limit.

On 50 held-out test seeds, the recurrent policy achieves
$75\%$ coverage and $50\%$ completion, compared with $65\%$ and $12\%$ for the four-frame policy
(Table~\ref{tab:rl-memory}). Replacing the critic's separate LSTM with the policy's shared recurrent representation reduces coverage to approximately $61\%$. These comparisons support recurrent memory and separate policy and critic recurrence for the evaluated policies. However, the reported intervals describe variation across test environments, not independently
trained policies. Robustness to training randomness and the persistence of these gains on larger maps remain topics for future research.

\begin{table}[t]
    \centering
    \small
    \setlength{\tabcolsep}{3pt}
    \caption{Memory comparison on 50 held-out test seeds using
    $20\times20$ open maps. Entries are percentages; $\pm$ values are
    \gls{ci} half-widths in percentage points.}
    \label{tab:rl-memory}
    \begin{tabular*}{\columnwidth}{
        @{\extracolsep{\fill}}lccc@{}
    }
        \toprule
        \textbf{Policy}
        & \textbf{Coverage}
        & \textbf{Recall}
        & \textbf{Success} \\
        \midrule
        Four-frame & $65\pm6$ & $66\pm8$ & $12\pm9$ \\
        Recurrent  & $75\pm7$ & $79\pm8$ & $50\pm14$ \\
        \bottomrule
    \end{tabular*}
\end{table}

\subsection{Five-controller comparison}
\label{subsec:comparisonlocalcontrollers}

Using the same $\mathcal E^{\mathrm{RLE}}$ environment and
evaluation protocol, we compare
nominal \gls{mpc}, fixed-tube \gls{tmpc}, pure \gls{rl}, shielded
\gls{rl} (\emph{Shield}), and shielding with a completion fallback
(\emph{Shield+C}) over 250 held-out seeds. Each controller operates one
robot with sensing radius $\rrobot=3$.

Pure \gls{rl}, Shield, and Shield+C use the same recurrent policy.
Shield applies the action check from
Section~\ref{subsubsec:rl-controller}. Shield+C permanently switches to
the fixed-tube \gls{tmpc} backup after 40 consecutive steps without a
newly observed cell, addressing prolonged lack of progress separately from individual unsafe-action overrides. Sensitivity to the fixed 40-step threshold remains a topic for future evaluation.

Decision time measures action selection only, excluding sensing and simulation, on one AMD EPYC 7B13 CPU core.

Table~\ref{tab:rl-controller-comparison} shows that pure \gls{rl} is
fastest but records $117.7$ collisions and $175.8$ harm
events per episode. Both shielded variants record zero terrain and victim collisions
and approximately $12$ harm events, a reduction of about $93\%$.
The remaining events arise from vulnerable-terrain exposure, which can
persist under imperfect perception and backup commands that are not
subject to the shield's vulnerability threshold. Shielding therefore
greatly reduces the measured harm without guaranteeing zero exposure.

\begin{table*}[t]
    \centering
    \small
    \setlength{\tabcolsep}{3pt}
    \caption{Final comparison over 250 test seeds. Success, coverage,
    and victim recall are percentages; $\pm$ values are \gls{ci}
    half-widths in percentage points. Collisions and harm are mean
    counts per episode; time is mean milliseconds per decision.
    No terrain or victim collisions were observed for the shielded variants. \glspl{ci} use the normal approximation $\text{mean}\pm1.96\,\text{standard error}$ across test seeds.}
    \label{tab:rl-controller-comparison}
    \begin{tabular*}{\textwidth}{@{\extracolsep{\fill}}lrrrrrr@{}}
        \toprule
        \textbf{Controller} & \textbf{Success} & \textbf{Coverage}
        & \textbf{Recall} & \textbf{Collisions} & \textbf{Harm} & \textbf{Time (ms)} \\
        \midrule
        \gls{mpc} & $94\pm3$ & 79 & $95\pm3$ & 0.2 & 6.5 & 274 \\
        \gls{tmpc} & $36\pm6$ & 57 & $63\pm5$ & 0 & 11.9 & 129 \\
        Pure \gls{rl} & $43\pm6$ & 69 & $77\pm4$ & 117.7 & 175.8 & 12 \\
        Shield & $30\pm6$ & 64 & $71\pm4$ & 0 & 12.0 & 40 \\
        Shield+C & $40\pm6$ & 64 & $72\pm4$ & 0 & 12.1 & 53 \\
        \bottomrule
    \end{tabular*}
\end{table*}

Adding the completion fallback raises observed success from $30\%$ for Shield to $40\%$ for Shield+C. Shield replaces approximately $37\%$ of policy proposals with backup commands. For Shield+C, the backup supplies approximately $65\%$ of
commands, including both safety overrides and commands issued after permanent fallback.

Wilcoxon signed-rank tests compare controllers over matched test seeds. Compared with fixed-tube \gls{tmpc}, Shield+C
improves coverage and recall by 7 and 9 percentage points
and reduces decision time from 129 to 53~ms
(all $p<0.01$). Completion differences from pure \gls{rl}
($p=0.42$) and fixed-tube \gls{tmpc} ($p=0.18$)
are not statistically significant.

The fallback does not guarantee completion: unsuccessful
episodes reach the detection deadline, and the
results do not identify whether stalled exploration,
backup infeasibility, or evolving fire prevents completion.

Nominal \gls{mpc} achieves the highest completion ($94\%$) and lowest
mean harm (6.5), but requires 274~ms per decision. The absence of tube tightening may benefit nominal \gls{mpc}
in the wide, non-vulnerable corridors, although its fire forecast
also differs from that of the tube-based controller. These effects
were not isolated experimentally. 

\begin{figure}[t]
    \centering
    \includegraphics[width=0.8\linewidth]{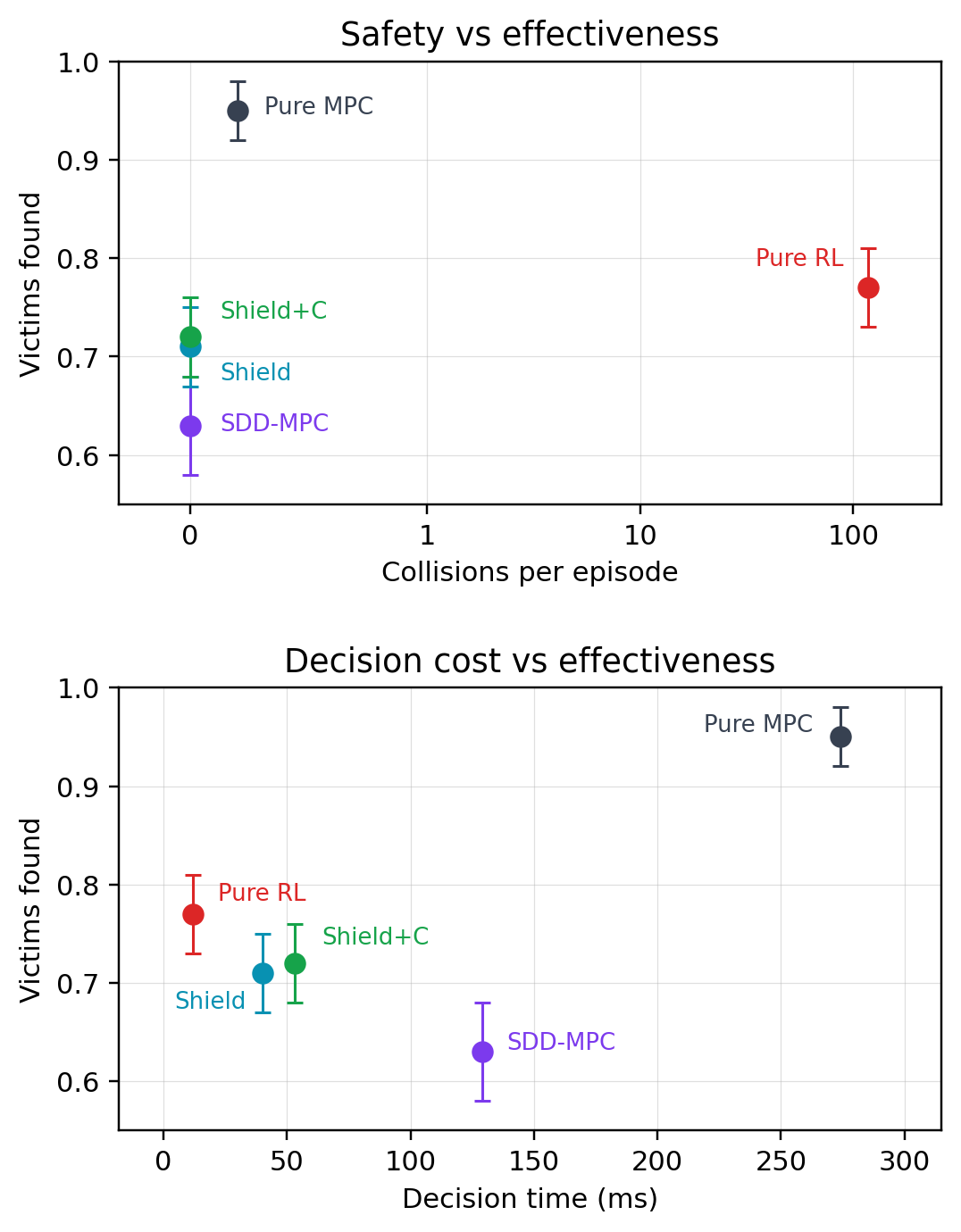}
    \caption{Mission effectiveness versus measured harm and decision
    time in the five-controller comparison. Error bars denote \glspl{ci}. Controllers are compared within the same evaluation setting.}
    \label{fig:rl-controller-tradeoffs}
\end{figure}

Figure~\ref{fig:rl-controller-tradeoffs} summarizes the
architectural trade-off: combining learned proposals with
a perception-based shield and model-based backup provides
an intermediate option between fast unshielded exploration and the higher completion of nominal \gls{mpc}.

\subsection{Supervisory coordination}% ($\esr$)}
\label{subsec:supervisory-rescue-experiments}

We first evaluate how shared perception and supervisory task allocation
affect multi-robot search, and subsequently examine heterogeneous
exploration and rescue. For team studies, coverage is the fraction of traversable cells observed by at least one robot. Exploration velocity is this
team-wide coverage divided by elapsed mission steps. Search distribution is further
characterized through inter-robot separation and normalized exploration
redundancy. At a given step, let $N_r^{\mathrm{observed}}$ denote the number
of distinct traversable cells observed by robot $r$ since
initialization, and let $N^{\mathrm{global}}$ denote
the number observed by at least one team member. For $|\mathcal R|>1$ and $N^{\mathrm{global}}>0$, normalized
exploration redundancy is defined as

\begin{equation}
\label{eqn:exploration-redundancy}
D^{\mathrm{redundancy}}
=
\dfrac{
\sum_{r\in\mathcal R}N_r^{\mathrm{observed}}-N^{\mathrm{global}}
}{
N^{\mathrm{global}}(|\mathcal R|-1)
}.
\end{equation}

Zero denotes disjoint coverage and one complete overlap. The measure captures overlap between robots' cumulative
observations.

Mean inter-robot distance is averaged across robot pairs and
simulation steps. Cumulative vulnerable-terrain exposure counts
team-wide robot-steps on cells with non-zero ground-truth
vulnerability. Harm events follow the definition in
Section~\ref{subsec:common-experimental-setup}. Pairwise comparisons of supervisory architectures use
two-sided paired $t$-tests on matched seeds with available values for both architectures. Reported $p$-values are not adjusted for multiple comparisons.

\subsubsection{Search efficiency and mission safety}
\label{subsec:supervisorsearchefficiency}

The coordination study compares three four-robot teams over 216 matched
$60\times60$ maps with 20 stationary victims, one initial fire, and a
300-step limit. The supervised team uses the shared belief and event-driven
allocation of Section~\ref{subsec:supervisorycoordination}.
The unsupervised team uses independent local perceptions without
central allocation. The shared team combines team observations without allocating tasks. All
robots use task-aware nominal \gls{mpc} and explore in idle mode when
unassigned. Because the study only evaluates victim detection, all robots have zero rescue capability and the supervisor does not generate rescue tasks. A run terminates upon detection completion, as defined in
Section~\ref{subsec:common-experimental-setup}, or when the 300-step limit is reached.

Shared information accounts for much of the coverage improvement. The mean time to observe $70\%$ of traversable cells is
108 steps for both the supervised and shared teams, compared with
137 steps for the unsupervised team. For $90\%$ coverage, the
corresponding means are 151, 153, and 201 steps
(Table~\ref{tab:sr-exploration-milestones},
Appendix~\ref{app:supplementary-results}).
The supervised team also increases mean inter-robot separation over the unsupervised team
(19.69 versus 18.02 cells, $p<0.001$), but no difference from the shared team is
detected ($p=0.139$). Information sharing therefore distributes robots
effectively even without explicit assignments; coverage milestones alone
show only a small additional supervisory benefit.

Allocation adds value by reducing overlap and improving task completion.
Normalized redundancy \eqref{eqn:exploration-redundancy} is $0.355$ for the supervised team, $0.461$ for the shared team,
and $0.623$ for the unsupervised team. Exploration velocity rises from $0.378\%$ per
step for the unsupervised team and $0.495\%$ for the shared team to $0.549\%$ for
the supervised team (both comparisons $p<0.001$): lower spatial overlap accompanies greater coverage gained per step.

\begin{figure*}[t]
    \centering
    \includegraphics[width=\textwidth]{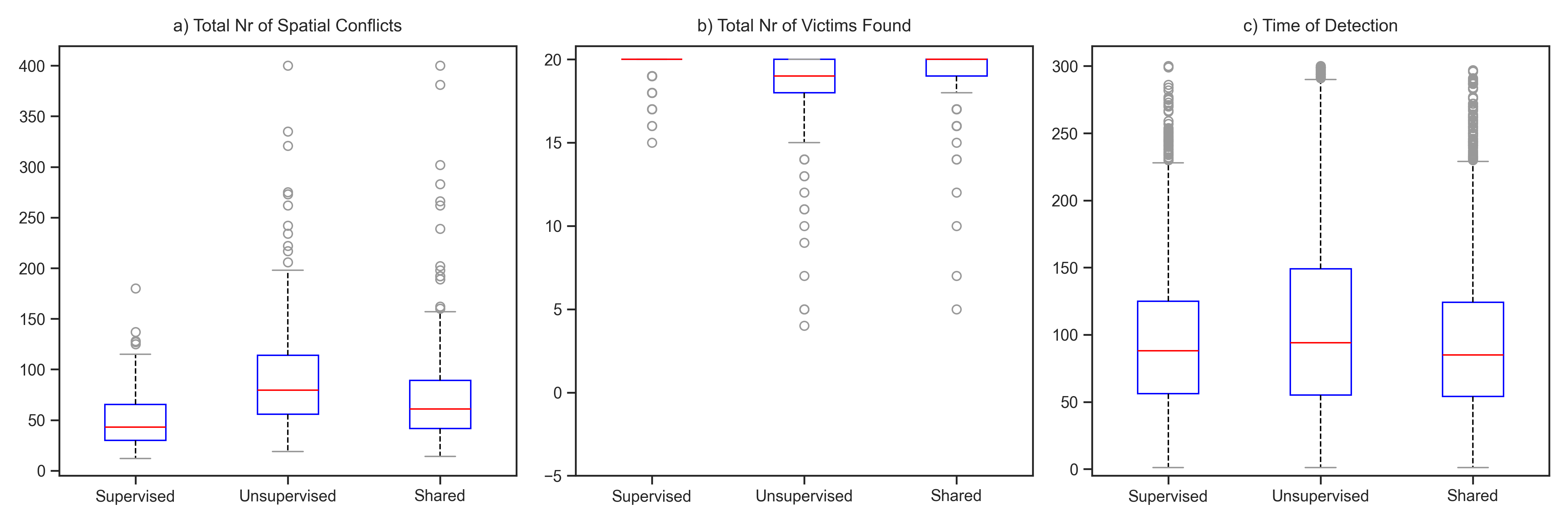}
    \caption{Coordination-study distributions:
    (a) spatial-conflict counts per run;
    (b) detected-victim counts at run termination; and
    (c) detection-update times pooled across runs.
    Boxes show the interquartile range and red lines the median;
    whiskers extend to the most extreme observations within
    $1.5$ interquartile ranges of the box.
    Panels (a) and (b) contain one observation per run.
    In panel (c), each observation is a step at which the logged
    detected-victim count increases; a run may contribute several
    timestamps, while simultaneous detections contribute one.}
    \label{fig:sr-coordination-distributions}
\end{figure*}

The mission-level effect becomes clearer in
Table~\ref{tab:sr-victim-discovery}: the supervised team achieves detection completion in $91.7\%$ of runs, compared with $75.9\%$ for the shared team
and $50.5\%$ for the unsupervised team. It also has the lowest mean discovery time and standard deviation among each architecture's successful
runs, although these subsets need not contain the same seeds. Figure~\ref{fig:sr-coordination-distributions} shows the distributions of spatial conflicts, detected-victim counts, and detection-update times.

\begin{table}[t]
    \centering
    \small
    \caption{Victim-discovery performance over 216 matched seeds. Detection completion requires all surviving victims to have been detected, with at least one survivor. Victims removed by fire are excluded from this condition. Completion times are reported only for runs meeting that condition. Duration is the mean $\pm$ standard deviation among each architecture's successful runs;
    these subsets need not contain the same seeds.}
    \label{tab:sr-victim-discovery}
    \begin{tabular*}{\columnwidth}{
        @{\extracolsep{\fill}}lcc@{}
    }
        \toprule
        \textbf{Architecture}
        & \makecell{\textbf{Detection}\\\textbf{completion (\%)}}
        & \textbf{Duration (steps)} \\
        \midrule
        Supervised team   & 91.7 & $177.59\pm43.93$ \\
        Unsupervised team & 50.5 & $220.81\pm53.37$ \\
        Shared team      & 75.9 & $184.02\pm50.03$ \\
        \bottomrule
    \end{tabular*}
\end{table}

The supervised team reduces mean inter-robot collisions from 15.58 to 0.99 per mission relative to the unsupervised team ($94\%$, $p=0.008$), and total harm
events by $25\%$ ($p<0.001$). Relative to the shared team, harm falls by $6\%$ ($p=0.039$); the $86\%$ lower inter-robot collision count is
not statistically conclusive ($p=0.068$). These observations are
consistent with spatially separated assignments reducing encounters,
although the experiment does not isolate each supervisory mechanism.
Coordination can therefore improve search efficiency without increasing
the recorded aggregate harm.

The supervised team also accumulates fewer vulnerable-terrain robot-steps than
the unsupervised team ($286.68$ versus $364.29$), despite a higher exposed fraction
of its shorter missions. The distinction matters operationally: cumulative exposure decreases even when the proportion of time spent in vulnerable cells increases.

Reported mean simulation-step times are similar for the supervised team and the unsupervised team (1.076 and 1.084~s, $p=0.338$). With 50.6 interventions
per mission and 11~ms per intervention on average, allocation adds little measured overhead relative to local simulation and planning.

The increased-fire comparison evaluates scenarios with 1, 3, 5, and 10 initial fire points. Increasing the initial fire count from one to ten reduces reported detection completion from $91.7\%$ to $54.6\%$ for
Supervised, $75.9\%$ to $44.4\%$ for the shared team, and $50.5\%$ to
$31.9\%$ for the unsupervised team. The supervised team retains the highest rate at every tested fire count and $92$--$94\%$ fewer inter-robot
collisions than the unsupervised team. However, fire both obstructs search and removes victims from the detection-completion condition. These results therefore compare performance under increasing fire, but do not isolate navigation difficulty or
establish detection of all initially placed victims.

The comparison distinguishes the roles of information sharing
and task allocation within \gls{pasar}: shared observations
improve collective coverage, while supervisory assignments
further reduce duplicated search and increase detection
completion.

\subsubsection{Heterogeneous-team performance}
\label{subsec:supervisorheterogenousteam}
The heterogeneous comparison evaluates the supervised and
unsupervised teams over 100 matched runs on $50\times50$
maps with 10 victims, one initial fire, and a 300-step
evaluation horizon. Each team contains two nominal-\gls{mpc} explorers, one \gls{sddtmpc} explorer, and one rescue-capable robot using task-aware nominal \gls{mpc} before active guidance. All four robots can explore, but only the rescue-capable robot can accept rescue tasks. Moreover, only its scans create victim-tracker entries and initiate rescue allocation; detections
by the other robots contribute to the team detection metric but do not create rescue tasks.

Here, guidance starts when the robot occupies its stored
rescue-task target or a nonempty \gls{bfs} path to that target
contains at most three vertices, including its start and
target. The path is recomputed from the robot's local perception.
Victims remain stationary until they perceive the rescue-capable
robot, after which they follow it immediately. This replaces
the trust-dependent following decision of
Section~\ref{subsec:victims}, separating the allocation comparison
from trust-dependent victim responses.

The supervised team reaches the $50\%$ and $90\%$ coverage milestones in fewer
steps and achieves a higher exploration velocity than the unsupervised team
(all $p<0.001$). It also increases mean inter-robot separation from
$20.45$ to $21.82$ cells ($p<0.001$). The exploration benefit of
supervision therefore persists when robots have heterogeneous
controllers and capabilities.

Despite its exploration gains, the supervised team rescues fewer
victims than the unsupervised team ($2.21$ versus $2.83$,
$p<0.001$). By representing exploration and rescue within the
same architecture, \gls{pasar} exposes this mismatch between
search performance and evacuation outcomes, motivating the
following studies of rescue-team size and adaptive guidance.
Future work could vary rescue capacity within the supervisory
comparison and incorporate rescue workload into task allocation
to better translate discoveries into completed evacuations.

\subsection{Rescue study and adaptive guidance}
\label{subsec:rescueandguidance}

The rescue study uses $45\times45$ maps with 10 dynamic victims and one
initial fire. Unlike the immediate-following
configuration of the heterogeneous-team study, victim movement and
following are governed by the perception- and trust-dependent behavioral
model of Section~\ref{subsec:victims}. The same 100 seeds are evaluated
with $\lvert\mathcal R\rvert\in\{0,2,4,6,8\}$ rescue robots initialized
at the entrance. A run terminates when all victims have either been rescued or killed, or when the 300-step limit is reached. Confidence intervals for the mean outcomes at each team
size use the normal approximation
$\text{mean}\pm1.96\,\text{standard error}$ across seeds. Averaging quadratic fits to the number of rescued victims obtained for each seed gives

\begin{equation}
\label{eqn:rescue-team-regression}
\widehat N^{\mathrm{resc}}
=1.899+0.440\abs{\mathcal R}-0.031\abs{\mathcal R}^2.
\end{equation}

\begin{figure}[t]
    \centering
    \includegraphics[width=\linewidth]{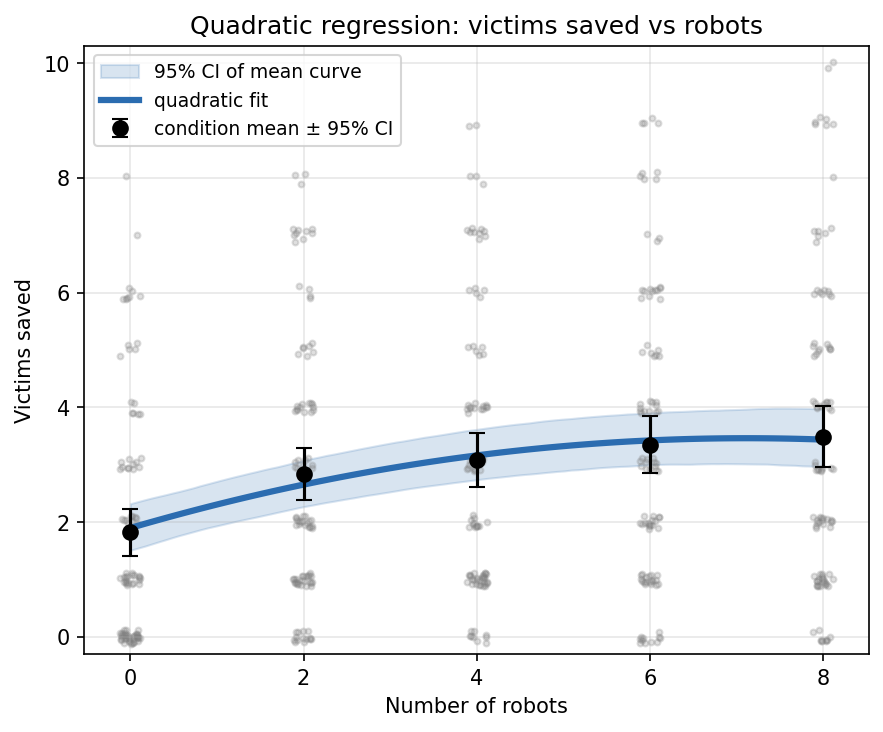}
    \caption{Rescue outcomes over 100 matched seeds per team size.
    Points and error bars show condition means and \glspl{ci}; the curve is the fitted quadratic.}
    \label{fig:sr-team-size}
\end{figure}

The negative quadratic coefficient ($p<0.001$) indicates diminishing returns: adding successive pairs of robots predicts approximately
$0.76$, $0.51$, $0.26$, and $0.01$ additional rescues per run. Increasing the team from four to six robots raises mean rescues
from 3.08 to 3.35 (Figure~\ref{fig:sr-team-size}). Four robots are selected as a resource--performance compromise
for the guidance comparison. Within \gls{pasar}, the result illustrates how explicit
rescue outcomes can inform team composition beyond maximizing
search capacity.

Using four robots and 100 matched seeds, the guidance ablation compares the adaptive policy from
Section~\ref{subsec:supervisorycoordination} with an always-lead policy
that ignores the guidance score \eqref{eqn:robot-trust-estimate}. Outcome and trust diagnostics are compared using two-sided
paired $t$-tests. Confidence intervals for mean paired
differences use 5000 percentile-bootstrap resamples of
the run-level differences. 

Adaptive guidance rescues $3.08$ compared to $2.34$ victims per run for the always-lead policy: a paired gain of $0.74$ (\gls{ci} $[0.51,0.96]$,
$p<0.001$). Adaptive guidance leaves $0.66$ fewer victims remaining in the environment
(\gls{ci} $[-0.89,-0.42]$, $p<0.001$), while no mortality difference is
detected ($p=0.402$). Table~\ref{tab:sr-guidance-outcomes} reports the complete numerical results.

The results show no detected increase in final trust
or mean trust during regular updates. Adaptive guidance produces
more decreases within regular trust updates, excluding
timeout-induced resets
(Table~\ref{tab:sr-guidance-trust}). The evacuation gain therefore
cannot be attributed to an observed increase in these trust
measures. More favorable robot--victim positioning is a possible explanation, but was not directly measured.

By representing victim responses separately from robot guidance,
\gls{pasar} supports evaluation of both rescue-team composition
and robot--victim interaction through evacuation outcomes.
The diminishing returns from additional rescuers and the gains
from adaptive guidance with a fixed team show that rescue
performance depends on both team size and guidance policy.

\glsresetall

\section{Conclusion}
\label{sec:conclusion}

This paper presented \gls{pasar} and evaluated how local
control, collective search, and victim guidance contribute
to search and evacuation outcomes.

For local control, \gls{sddtmpc} completes more missions than
fixed-tube \gls{tmpc} while retaining the observed protection
against terrain and victim collisions under noise-free conditions.
State-dependent tightening therefore offers a practical balance
between exploration and disturbance protection, although larger
perception errors and spreading fire remain challenges.
For learned exploration, the evaluated recurrent policy
achieves higher coverage and completion than the four-frame policy. Shielding the learned
policy eliminates recorded terrain and victim collisions while
retaining shorter decision times than the model-based baselines.
Nominal \gls{mpc} nevertheless achieves higher completion in the
tested open maps, demonstrating the trade-off between decision
speed, collision avoidance, and mission completion.

The coordination studies distinguish the benefits of shared
perception from those of task allocation. Sharing observations
improves collective search, while supervisory allocation further
reduces redundant exploration and increases detection completion
under the study's remaining-victim endpoint. In the heterogeneous
configuration, however, exploration gains accompany fewer rescues.
By representing search and rescue as coupled responsibilities,
\gls{pasar} exposes this mismatch and motivates allocation
strategies that account for evacuation capacity.

The dynamic-rescue studies show diminishing returns from additional
rescuers and improved evacuation through adaptive approach--lead
guidance with a fixed team. The guidance benefit appears in the
number of completed rescues, without a detected increase in logged
trust or reduction in mortality. Together, these findings demonstrate
the importance of evaluating robot--victim interaction alongside
team composition and victim discovery.

Future work should evaluate more complete configurations combining
local control, supervisory allocation, and adaptive guidance.
Priorities include uncertainty-aware constraints for evolving
hazards, independent variation of sensing and fire-update rates,
and learned-policy evaluation across training seeds and map
distributions. Rescue-workload-aware allocation and measurements
of robot--victim separation and guidance interruptions would
clarify how search gains translate into evacuation. Physical validation is a further step toward establishing
performance beyond the simulated environments and behavioral
models considered here. 

Together, these studies establish the value of \gls{pasar}
as a framework for identifying where improvements in local
control and collective search support evacuation, and where
they fall short. By making these dependencies explicit,
\gls{pasar} provides a basis for selecting controllers,
allocating rescue capacity, and designing guidance policies
according to mission outcomes rather than isolated performance
measures.

\paragraph{Data availability statement}

The source code supporting this study is available at
\url{https://github.com/svenschoonebeek/Perception-Aware-Search-and-Rescue}.

\bibliography{sn-bibliography}% common bib file

\onecolumn

\begin{appendices}

\section{Supplementary Experimental Results}
\label{app:supplementary-results}

\setcounter{table}{0}
\renewcommand{\thetable}{A\arabic{table}}

\captionsetup[table]{
    justification=raggedright,
    singlelinecheck=false
}

\par\noindent
\begin{minipage}{\linewidth}
\centering
\setlength{\tabcolsep}{2.5pt}
\captionof{table}{Mean steps to reach coverage milestones in the
216-seed coordination study, rounded to whole steps. Each mean includes only runs reaching the corresponding milestone, and sample sizes may differ across milestones and architectures.}
\label{tab:sr-exploration-milestones}

\begin{tabular*}{\linewidth}{
    @{\extracolsep{\fill}}lccccc@{}
}
    \toprule
    \textbf{Architecture}
    & \textbf{50\%}
    & \textbf{70\%}
    & \textbf{80\%}
    & \textbf{85\%}
    & \textbf{90\%} \\
    \midrule
    Supervised team  & 81 & 108 & 126 & 136 & 151 \\
    Unsupervised team & 92 & 137 & 165 & 183 & 201 \\
    Shared team     & 80 & 108 & 128 & 139 & 153 \\
    \bottomrule
\end{tabular*}

\end{minipage}
\par\medskip

\par\noindent
\begin{minipage}{\linewidth}
\centering
\setlength{\tabcolsep}{3pt}

\captionof{table}{Evacuation outcomes over 100 matched runs with 10 victims per run. Adaptive and always-lead entries are mean victim counts. Differences are adaptive minus always lead; \glspl{ci} refer to the paired mean differences.}
\label{tab:sr-guidance-outcomes}

\begin{tabular*}{\linewidth}{
    @{\extracolsep{\fill}}lccccc@{}
}
    \toprule
    \textbf{Metric}
    & \textbf{Adaptive}
    & \textbf{Always lead}
    & \textbf{Difference}
    & \textbf{\gls{ci}}
    & \textbf{$p$} \\
    \midrule
    Rescued
    & 3.08 & 2.34 & $+0.74$ & $[0.51,0.96]$ & $<0.001$ \\
    Deceased
    & 2.05 & 2.13 & $-0.08$ & $[-0.27,0.10]$ & $0.402$ \\
    Remaining
    & 4.87 & 5.53 & $-0.66$ & $[-0.89,-0.42]$ & $<0.001$ \\
    \bottomrule
\end{tabular*}

\end{minipage}
\par\medskip

\noindent
\begin{minipage}{\linewidth}
\centering
\setlength{\tabcolsep}{3pt}

\captionof{table}{Victim-trust and following diagnostics over
100 matched runs. Final and mean trust are dimensionless;
trust drops are counts, and following is expressed as a
percentage. \glspl{ci} describe adaptive-minus-always-lead
differences, in percentage points for following.
Final trust is measured at victim removal or run termination.
Mean trust averages post-update values during regular updates,
after initialization and while a rescue robot is perceived;
trust-drop counts exclude timeout resets. The following ratio
divides selected following commands by logged trust-update
steps. Metrics are computed per victim, averaged within each
run, and then across runs. Victims without logged updates are
excluded from mean-trust and following-ratio calculations.}
\label{tab:sr-guidance-trust}

\begin{tabular*}{\linewidth}{
    @{\extracolsep{\fill}}lcccc@{}
}
    \toprule
    \textbf{Metric}
    & \textbf{Adaptive guidance}
    & \textbf{Always lead}
    & \textbf{\gls{ci} of difference}
    & \textbf{$p$} \\
    \midrule
    Final trust
    & 0.503 & 0.500 & $[-0.017,0.025]$ & 0.724 \\
    Mean trust
    & 0.511 & 0.497 & $[-0.010,0.039]$ & 0.278 \\
    Trust drops
    & 1.074 & 0.237 & $[0.626,1.062]$ & $<0.001$ \\
    Following (\%)
    & 43.5 & 46.5 & $[-6.0,0.0]$ & 0.051 \\
    \bottomrule
\end{tabular*}

\end{minipage}
\par\medskip

\noindent
\begin{minipage}{\linewidth}
\centering
\small
\setlength{\tabcolsep}{4pt}

\captionof{table}{Frame-stacked \gls{rl} sensitivity study.
Each configuration represents one policy trained with seed zero, and
outcomes are averaged over its final training rollouts. Coverage, victim
recall, and success are reported as percentages.}
\label{tab:rl-sensitivity}

\begin{tabular*}{\linewidth}{
    @{\extracolsep{\fill}}lcccc@{}
}
    \toprule
    \textbf{Configuration}
    & \textbf{Map}
    & \textbf{Coverage (\%)}
    & \textbf{Victim recall (\%)}
    & \textbf{Success (\%)} \\
    \midrule
    Reference configuration
    & $20\times20$ & 60.5 & 57.2 & 14.0 \\
    $\gamma^{\mathrm{RL}}=0.997$
    & $20\times20$ & 61.5 & 59.7 & 17.0 \\
    $\gamma^{\mathrm{RL}}=0.999$
    & $20\times20$ & 62.1 & 60.5 & 16.0 \\
    Random start
    & $20\times20$ & 58.6 & 62.8 & 17.0 \\
    $\gamma^{\mathrm{RL}}=0.999$ and random start
    & $20\times20$ & 60.3 & 64.2 & 15.0 \\
    Episode horizon 1000
    & $20\times20$ & 61.7 & 65.2 & 26.0 \\
    Frontier potential
    & $20\times20$ & 59.3 & 60.5 & 18.0 \\
    Four million training steps
    & $20\times20$ & 56.2 & 61.7 & 21.0 \\
    Scale test
    & $25\times25$ & 53.1 & 55.0 & 4.0 \\
    Scale test
    & $30\times30$ & 41.5 & 41.0 & 0.0 \\
    \bottomrule
\end{tabular*}

\end{minipage}
\par\medskip

\clearpage

\section{Experimental Configurations and Model Parameters}
\label{app:configurations}

\setcounter{table}{0}
\renewcommand{\thetable}{B\arabic{table}}

\begingroup
\small
\setlength{\tabcolsep}{3pt}
\renewcommand{\arraystretch}{1.15}
\begin{longtable}{@{}
>{\raggedright\arraybackslash}p{\dimexpr0.18\textwidth-1.5\tabcolsep\relax}
>{\raggedright\arraybackslash}p{\dimexpr0.35\textwidth-1.5\tabcolsep\relax}
>{\raggedright\arraybackslash}p{\dimexpr0.14\textwidth-1.5\tabcolsep\relax}
>{\raggedright\arraybackslash}p{\dimexpr0.33\textwidth-1.5\tabcolsep\relax}
@{}}
\caption[Experimental configurations and model parameters.]{
Experimental configurations and model parameters.
Scenario notation follows Section~\ref{sec:experiments};
$\mathrm{hom}$ and $\mathrm{het}$ distinguish homogeneous
and heterogeneous supervisory teams. Trust-update and autonomous-movement parameters in Block E
apply to $\mathcal E^{\mathrm{DR}}$, whereas scenario-specific sensing and guidance settings are identified separately.}
\label{tab:model-variables}\\
\toprule
\textbf{Parameter/setting} & \textbf{Meaning and scope}
& \textbf{Reference} & \textbf{Value/configuration}\\
\midrule
\endfirsthead
\multicolumn{4}{c}{\tablename\ \thetable\ (continued)}\\
\toprule
\textbf{Parameter/setting} & \textbf{Meaning and scope}
& \textbf{Reference} & \textbf{Value/configuration}\\
\midrule
\endhead
\midrule
\multicolumn{4}{r}{Continued on next page}\\
\endfoot
\bottomrule
\endlastfoot

\midrule
\multicolumn{4}{@{}l}{\textbf{A. Environment generation}}\\*
\midrule
Room count
& Number of generated rooms.
& --
& $\mathcal E^{\mathrm{MPC}}$: $7$;
$\mathcal E_{\mathrm{hom}}^{\mathrm{SV}}$: $12$;
$\mathcal E_{\mathrm{het}}^{\mathrm{SV}}$: $15$;
$\mathcal E^{\mathrm{DR}}$: $10$;
$\mathcal E^{\mathrm{RLD}}$: $7,10,14$ at sizes $20,25,30$;
$\mathcal E^{\mathrm{RLE}}$: $6$. \\
Room side bounds
& Configured minimum/maximum width and length, in cells.
& --
& $\mathcal E^{\mathrm{MPC}}$: $(6,12)$;
$\mathcal E^{\mathrm{SV}}$: $(4,8)$;
$\mathcal E^{\mathrm{DR}}$: $(5,10)$;
$\mathcal E^{\mathrm{RLD}}$: $(3,5)$; $\mathcal E^{\mathrm{RLE}}$: $(3,6)$. \\
Corridor-width bounds
& Configured minimum/maximum corridor thickness.
& --
& $\mathcal E^{\mathrm{MPC}}$: $(1,3)$;
$\mathcal E^{\mathrm{SV}}$: $(2,3)$;
$\mathcal E^{\mathrm{DR}}$: $(2,3)$;
$\mathcal E^{\mathrm{RLD}}$: $(1,1)$. \\
Entrance dimensions
& Starting-room width and length, in cells.
& --
& All scenarios: $(3,3)$. \\
Room vulnerability probability
& Probability that a room receives non-zero vulnerability.
& --
& $\mathcal E^{\mathrm{MPC}}$: $0.3$;
$\mathcal E^{\mathrm{SV}}$: $0.3$;
$\mathcal E^{\mathrm{DR}}$: $0.4$;
$\mathcal E^{\mathrm{RLD}}$: $0.12$;
$\mathcal E^{\mathrm{RLE}}$: $0.35$. \\
Room high-risk probability
& Conditional probability of value 1 for a vulnerable room.
& --
& $\mathcal E^{\mathrm{MPC}}$: $0.4$;
$\mathcal E^{\mathrm{SV}}$: $0.2$;
$\mathcal E^{\mathrm{DR}}$: $0.2$;
$\mathcal E^{\mathrm{RLD}}$: $0.3$;
$\mathcal E^{\mathrm{RLE}}$: $0.25$. \\
Corridor vulnerability probability
& Probability that a corridor receives non-zero vulnerability.
& --
& $\mathcal E^{\mathrm{MPC}}$: $0.3$;
$\mathcal E^{\mathrm{SV}}$: $0.4$;
$\mathcal E^{\mathrm{DR}}$: $0.1$;
$\mathcal E^{\mathrm{RLD}}$: $0.08$;
$\mathcal E^{\mathrm{RLE}}$: $0$. \\
Corridor high-risk probability
& Conditional probability of value 1 for a vulnerable corridor.
& --
& $\mathcal E^{\mathrm{MPC}}$: $0.4$;
$\mathcal E^{\mathrm{SV}}$: $0.2$;
$\mathcal E^{\mathrm{DR}}$: $0.4$;
$\mathcal E^{\mathrm{RLD}}$: $0.3$. \\

\midrule
\multicolumn{4}{@{}l}{\textbf{B. Sensing, fire, and displacement models}}\\*
\midrule
$\eta_r^0$
& Base scan quality of robot $r$.
& \eqref{eqn:scanquality}
& $\mathcal E^{\mathrm{MPC}}$: $0.35$;
$\mathcal E^{\mathrm{SV}}$, $\mathcal E^{\mathrm{DR}}$,
$\mathcal E^{\mathrm{RLD}}$, and $\mathcal E^{\mathrm{RLE}}$:
$0.9$. \\
$\rrobot$
& Euclidean sensing radius of robot $r$, in cells.
& \eqref{eqn:scanquality}
& $\mathcal E^{\mathrm{MPC}}$, $\mathcal E^{\mathrm{SV}}$,
and $\mathcal E^{\mathrm{DR}}$: $4$;
$\mathcal E^{\mathrm{RLD}}$: $3$. \\
Distance falloff
& Gaussian distance dependence of scan quality.
& \eqref{eqn:scanquality}
& $\mathcal E^{\mathrm{MPC}}$, $\mathcal E^{\mathrm{RLD}}$,
and $\mathcal E^{\mathrm{RLE}}$: enabled;
$\mathcal E^{\mathrm{SV}}$ and $\mathcal E^{\mathrm{DR}}$:
disabled. \\
$\deltavuln$
& Half-width of the uniform vulnerability-measurement error.
& --
& $\mathcal E^{\mathrm{SV}}$, $\mathcal E^{\mathrm{DR}}$,
$\mathcal E^{\mathrm{RLD}}$, and $\mathcal E^{\mathrm{RLE}}$:
$0$. \\
$\sigmaaconf$
& Per-step decay of confidence for an unobserved cell.
& \eqref{eqn:pconfijupdate}
& $\mathcal E^{\mathrm{MPC}}$: $0.05$;
$\mathcal E^{\mathrm{SV}}$, $\mathcal E^{\mathrm{DR}}$,
$\mathcal E^{\mathrm{RLD}}$, and $\mathcal E^{\mathrm{RLE}}$:
$0.01$. \\
$\pignite$
& Ground-truth ignition probability per burning--flammable
adjacency and fire update.
& --
& $\mathcal E^{\mathrm{MPC}}$: $0.04$;
$\mathcal E^{\mathrm{SV}}$: $0.01$;
$\mathcal E^{\mathrm{DR}}$: $0.02$;
$\mathcal E^{\mathrm{RLD}}$ and $\mathcal E^{\mathrm{RLE}}$:
$0.03$. \\
$T^{\mathrm{burn}}$
& Ground-truth burning duration in fire-update steps.
& --
& $\mathcal E^{\mathrm{MPC}}$: $14$;
$\mathcal E^{\mathrm{SV}}$ and $\mathcal E^{\mathrm{DR}}$:
$100$;
$\mathcal E^{\mathrm{RLD}}$ and $\mathcal E^{\mathrm{RLE}}$:
$8$. \\
$\pignitehat$
& Ignition probability assumed by the MPC fire predictor.
& --
& $\mathcal E^{\mathrm{MPC}}$: $1$;
$\mathcal E^{\mathrm{SV}}$ and $\mathcal E^{\mathrm{DR}}$: $0.3$;
$\mathcal E^{\mathrm{RLE}}$: $0.3$ for nominal MPC and $1$ for TMPC, Shield, and Shield+C. \\
$\tburnhat$
& Burning-age threshold used by the controller-side fire predictor.
& --
& $\mathcal E^{\mathrm{MPC}}$ and $\mathcal E^{\mathrm{SV}}$:
$10$ counter increments.
The MPC predictors in $\mathcal E^{\mathrm{DR}}$ and
$\mathcal E^{\mathrm{RLE}}$ omit burnout prediction. \\
$\alphatrip$
& Confidence scaling in the displacement probability.
& \eqref{eqn:slip-chance}
& $\mathcal E^{\mathrm{MPC}}$, $\mathcal E^{\mathrm{SV}}$,
and $\mathcal E^{\mathrm{DR}}$: $1$;
$\mathcal E^{\mathrm{RLD}}$ and $\mathcal E^{\mathrm{RLE}}$:
$0.7$. \\

$\betatrip$
& Exponent in the displacement probability.
& \eqref{eqn:slip-chance}
& $\mathcal E^{\mathrm{MPC}}$, $\mathcal E^{\mathrm{SV}}$,
and $\mathcal E^{\mathrm{DR}}$: $2$;
$\mathcal E^{\mathrm{RLD}}$ and $\mathcal E^{\mathrm{RLE}}$:
$1$. \\

\midrule
\multicolumn{4}{@{}l}{\textbf{C. MPC objectives and task-aware local control}}\\*
\midrule
Unreachable-frontier fallback
& Assigned value of the normalized frontier distance
$d_r^{\mathrm{front}}$ when no unexplored cell is reachable.
& \eqref{eqn:mpc-terminal-objective}
& $-1/(\nh+\nw)$. \\
$N$
& \gls{mpc} prediction horizon.
& \eqref{eqn:mpc-objective}
& $4$ \\
$\gamma^{\mathrm{MPC}}$
& \gls{mpc} discount factor.
& \eqref{eqn:mpc-objective}
& $0.9$ \\
$\wdisc$
& Discovery weight in the common \gls{mpc} stage objective. 
& \eqref{eqn:mpc-stage-objective}
& $100$ \\
$\wconf$
& Confidence weight in the common \gls{mpc} stage objective. 
& \eqref{eqn:mpc-stage-objective}
& $3$ \\
$\wterm$
& Terminal frontier-distance weight.
& \eqref{eqn:mpc-terminal-objective}
& $300$ \\
$\wvuln$
& Vulnerability weight in the common \gls{mpc} stage objective.
& \eqref{eqn:mpc-stage-objective}
& $\mathcal E^{\mathrm{MPC}}$: $0$;
$\mathcal E^{\mathrm{RLE}}$: $10$. \\
$\omegadisc$
& Discovery weight for an idle supervised robot.
& \eqref{eqn:task-aware-stage-objective}
& $100$ \\
$\omegaconf$
& Confidence weight for an idle supervised robot.
& \eqref{eqn:task-aware-stage-objective}
& $3$ \\
$\omegavuln$
& Vulnerability weight for an idle supervised robot.
& \eqref{eqn:task-aware-stage-objective}
& $10$ \\
$\omegaprox$
& robot-proximity weight for an idle supervised robot.
& \eqref{eqn:task-aware-stage-objective}
& $50$ \\
$\omegapath$
& Path-progress weight while following an assigned path.
& \eqref{eqn:task-aware-stage-objective}
& $100$ \\
$\omegaconftilde$
& Confidence weight while following an assigned path.
& \eqref{eqn:task-aware-stage-objective}
& $1$ \\
$\omegavulntilde$
& Vulnerability weight while following an assigned path.
& \eqref{eqn:task-aware-stage-objective}
& $5$ \\
$\omegaproxtilde$
& robot-proximity weight while following an assigned path.
& \eqref{eqn:task-aware-stage-objective}
& $200$ \\
$\beta_0^{\mathrm{path}},\beta_1^{\mathrm{path}},\beta_2^{\mathrm{path}}$
& Baseline, forward-progress, and path-deviation coefficients of the path score.
& Table~\ref{tab:mpc-objective-components-prox-path}
& $0.5,\ 0.1,\ 0.05$ \\
$n^{\mathrm{path}}$
& Number of remaining path cells considered when aligning the stored path.
& Table~\ref{tab:mpc-objective-components-prox-path}
& $10$ \\
$\psi(d)$
& Proximity penalty at Euclidean robot separation $d$.
& Table~\ref{tab:mpc-objective-components-prox-path}
& $1$ for $0<d\leq1.5$; $0.2$ for $1.5<d\leq2.5$; $0$ otherwise \\

\midrule
\multicolumn{4}{@{}l}{\textbf{D. Supervisory architecture and task allocation}}\\*
\midrule
$C_r^{\mathrm{explore}},C_r^{\mathrm{rescue}}$
& Auction coefficients determining eligibility and reward scaling for
exploration and rescue tasks.
& \eqref{eqn:supervisory-utility}
& $\mathcal E_{\mathrm{hom}}^{\mathrm{SV}}$: $(1,0)$ for all
robots; $\mathcal E_{\mathrm{het}}^{\mathrm{SV}}$: $(1,0)$ for the three
exploration-only robots and $(1,1)$ for the rescue-capable robot; $\mathcal E^{\mathrm{DR}}$: $(1,1)$ for all robots. \\
$\sigma^{\mathrm{sup}}$
& Per-step decay of unrefreshed supervisory confidence.
& --
& $\mathcal E^{\mathrm{SV}}$ and $\mathcal E^{\mathrm{DR}}$: $0.01$. \\
$B^{\mathrm{explore}},B^{\mathrm{rescue}}$
& Base rewards for exploration and rescue tasks.
& \eqref{eqn:supervisory-utility}
& $100,10000$ \\
$w^{\mathrm{task}}$
& Weight penalizing proximity to assigned exploration targets.
& \eqref{eqn:supervisory-spatial-penalty}
& $500$ \\
$w^{\mathrm{robot}}$
& Weight penalizing proximity to unavailable idle robots.
& \eqref{eqn:supervisory-spatial-penalty}
& $200$ \\
$\lambda^{\mathrm s}$
& Spatial decay length in cells of the supervisory redundancy penalty.
& \eqref{eqn:supervisory-spatial-penalty}
& $5$ \\
$n^{\mathrm{event}}$
& Number of pending noncritical events required to trigger intervention.
& --
& $2$ \\
$n^{\mathrm{impatience}}$
& Low-progress impatience threshold.
& --
& $8$ \\
$n^{\mathrm{conflict}}$
& Minimum interval between successive spatial-conflict
reports from the same robot.
& --
& $3$ \\

\midrule
\multicolumn{4}{@{}l}{\textbf{E. Dynamic victims and guidance}}\\*
\midrule
$\rvict$
& Euclidean sensing radius of a dynamic victim.
& --
& $\mathcal E^{\mathrm{SV}}:3$; $\mathcal E^{\mathrm{DR}}:4$. \\
Age assignment
& Initial age-group distribution.
& --
& Independent uniform draw from $\{\mathrm{child},\mathrm{adult},\mathrm{elderly}\}$ for each victim. \\
Interaction-strategy assignment
& Initial interaction-strategy distribution.
& --
& Independent uniform draw from altruistic, egoistic, and selfish for each victim. \\
$\theta_v(0)$
& Initial victim trust.
& \eqref{eqn:victim-trust-update}
& $0.5$ \\
$\delta_v^{\mathrm{trust}}$
& Trust sensitivity, fixed according to the victim's initial interaction
strategy.
& \eqref{eqn:victim-trust-update}
& $1.2$ if $\pi_v(0)=\mathrm{altruistic}$;
$1.0$ if $\pi_v(0)=\mathrm{egoistic}$;
$0.6$ if $\pi_v(0)=\mathrm{selfish}$. \\
$\theta^{\mathrm{follow}}$
& Victim trust threshold for robot-following behavior.
& --
& $0.4$ \\
$\wprox$
& Weight on positive reduction in victim--robot distance.
& \eqref{eqn:victim-trust-update}
& $0.05$ \\
$\wgain$
& Weight on non-negative change in perceived safety.
& \eqref{eqn:victim-trust-update}
& $0.10$ \\
$\wloss$
& Weight on negative change in perceived safety.
& \eqref{eqn:victim-trust-update}
& $0.30$ \\
$p^{\mathrm{move,} \kappa_v}$
& Base probability that a victim $v$ with age group $\kappa_v$ attempts to move.
& --
& child: $0.525$; adult: $0.90$; elderly: $0.65$ \\
$\mu^{\mathrm{danger}}$
& Multiplier applied to victim movement probability under perceived danger.
& --
& $1.5$; the resulting movement probability is capped at $1$. \\
$\sigma^{\mathrm{guide}}$
& Guidance-score update gain.
& \eqref{eqn:robot-trust-estimate}
& $0.1$ \\
$\varepsilon^{\mathrm{guide}}$
& Positive replacement for an unchanged robot--victim distance,
or an increase from a preceding separation within the
close-range threshold.
& \eqref{eqn:robot-trust-estimate}
& $0.01$ \\
Guidance-score close-range threshold
& Preceding robot--victim separation used to assign
$\varepsilon^{\mathrm{guide}}$ when separation increases.
& \eqref{eqn:robot-trust-estimate}
& $1.5$ cells. \\
$\theta^{\mathrm{lead}}$
& Guidance-score threshold separating approach and lead policies.
& --
& $0.4$ \\
$d^{\mathrm{init}}$
& Maximum Euclidean distance to the stored rescue-task target
for initiating guidance.
& --
& $\mathcal E^{\mathrm{DR}}$: $3$;
not used in $\mathcal E_{\mathrm{het}}^{\mathrm{SV}}$. \\

$d^{\mathrm{leash}}$
& Maximum robot--victim separation in cells during the lead policy.
& --
& $\mathcal E^{\mathrm{SV}}:5$; $\mathcal E^{\mathrm{DR}}:4$ \\
Guidance timeouts
& Consecutive leash-wait and no-progress steps before a rescue is abandoned.
& --
& $10$ and $15$, respectively \\
Victim stuck timeout
& Stationarity threshold for resetting the trust-update baseline.
& --
& $10$ steps. \\
$\mathcal C^{\mathrm{exit}}$
& Cells constituting the rescue-completion region.
& --
& $\{1,2,3\}\times\{1,2,3\}$. \\
$s_r^{\mathrm{guide}}(0)$
& Initial robot guidance score.
& \eqref{eqn:robot-trust-estimate}
& $0$ \\

\midrule
\multicolumn{4}{@{}l}{\textbf{F. $\mathcal E^{\mathrm{RLD}}$ and
$\mathcal E^{\mathrm{RLE}}$: final reward, observation, and shielding}}\\*
\midrule
$\wdet$
& Reward for each newly detected victim.
& \eqref{eqn:rl-reward}
& $10$ \\
$\wstep$
& Per-step \gls{rl} penalty.
& \eqref{eqn:rl-reward}
& $0.005$ \\
$\wterr$
& Penalty for a terrain collision.
& \eqref{eqn:rl-reward}
& $0.1$ \\
$\wvict$
& Penalty for a victim collision.
& \eqref{eqn:rl-reward}
& $1$ \\
$\wsucc$
& Reward for successful episode completion.
& \eqref{eqn:rl-reward}
& $50$ \\
$\wcov$
& Weight on terminal coverage.
& \eqref{eqn:rl-reward}
& $50$ \\
$\womega$
& Coverage-potential weight.
& \eqref{eqn:rl-reward}
& $30$ \\
$\gamma^{\mathrm{RL}}$
& \gls{rl} discount factor and potential-shaping discount.
& \eqref{eqn:rl-reward}
& $\mathcal E^{\mathrm{RLE}}$: $0.999$; \\
$\theta^{\mathrm{shield}}$
& Perceived vulnerability must be strictly below this threshold to pass the shield.
& --
& $0.5$ \\

\midrule
\multicolumn{4}{@{}l}{%
\textbf{G. $\mathcal E^{\mathrm{RLD}}$ and
$\mathcal E^{\mathrm{RLE}}$: experimental variants and training settings}}\\*
\midrule

$\mathcal E^{\mathrm{RLE}}$ recurrent-training configurations
& Seed and training-budget configurations specified in the
supplied training runner.
& --
& Seeds $0$, $1$, and $2$ with $2\times10^6$ training steps;
an additional seed-$0$ configuration with $4\times10^6$ steps.\\

Frame-stack reference configuration
& Reference configuration for the
$\mathcal E^{\mathrm{RLD}}$ sensitivity study.
& --
& 6 victims; $\gamma^{\mathrm{RL}}=0.99$; fixed initial position;
700-step episode limit; $2\times10^6$ training steps; training seed $0$. \\

Extended-episode variant
& Variant used to assess whether the episode limit restricts exploration.
& --
& $20\times20$ map; 6 victims;
$\gamma^{\mathrm{RL}}=0.999$; random initial position;
1000-step episode limit; $2\times10^6$ training steps. \\

Frontier-potential variant
& Alternative potential used in the shaping term of
\eqref{eqn:rl-reward}.
& --
& Negative distance to the nearest exploration frontier, normalized by
$\nh+\nw$, replaces the coverage potential. \\

Scale variants
& Configurations used to assess frame-stack performance on larger maps.
& --
& $25\times25$: 10 victims, 900-step episode limit, and
$3\times10^6$ training steps;
$30\times30$: 10 victims, 1200-step episode limit, and
$4\times10^6$ training steps. \\

Recurrent architecture
& Temporal representations evaluated in the memory study.
& --
& One 256-unit LSTM layer. The principal recurrent configuration also
uses a recurrent critic. \\

CNN encoder
& Shared spatial feature extractor.
& --
& Three convolutional layers with
$(32,64,64)$ channels, $3\times3$ kernels, strides $(1,2,2)$, and ReLU
activations, followed by a 256-dimensional feature representation. \\

Policy/value heads
& Hidden-layer widths of the policy and value heads.
& --
& $(256,256)$ for each head. \\

Learning rate
& --
& --
& Linear decay from $2.5\times10^{-4}$ to $0$. \\

Parallel environments
& --
& --
& $64$. \\

Rollout length
& Number of steps collected per environment before each update.
& --
& Frame-stack policy: $1024$;
recurrent policy: $512$. \\

Minibatch size and epochs
& --
& --
& 2048 samples and 4 epochs. \\

GAE parameter
& --
& --
& $\lambda=0.95$. \\

\gls{ppo} policy and value clipping
& --
& --
& $\epsilon=0.15$ and $\epsilon_V=0.2$. \\

Value-loss weight
& --
& --
& $c_V=0.5$. \\

Entropy coefficient
& Exploration-regularization schedule.
& --
& Annealed from $0.03$ to $0.005$. \\

Gradient clipping and KL target
& --
& --
& Maximum gradient norm $0.5$; target KL divergence $0.02$. \\

Normalization
& Reward, observation, and advantage normalization.
& --
& Reward normalization is enabled during training through
\texttt{VecNormalize}; observation normalization is disabled and PPO
advantage normalization is enabled. \\

\midrule
\multicolumn{4}{@{}l}{\textbf{H. Initialization}
}\\*
\midrule

Initial robot positions
& --
& --
& $\mathcal E^{\mathrm{MPC}}$: $(2,2)$;
$\mathcal E^{\mathrm{SV}}$: entrance region. \\

$M_{ij}^{\mathrm{burn}}(0)$
& Initial fire-age entry.
& \eqref{eqn:groundstate}
& $0$. \\

$\Pi_r(0)$
& Initial stored path.
& --
& Empty. \\

Initial unresolved-event queue
& --
& --
& $\varnothing$. \\

Victim history
& --
& --
& Initialized from the first applicable observation. \\
\end{longtable}
\endgroup

\end{appendices}

\end{document}